\documentclass[acmsmall]{acmart}
\AtBeginDocument{%
  }

\setcopyright{none}

\usepackage{amsthm}
\usepackage{subfig}
\usepackage{multirow}
\usepackage{makecell}
\usepackage[table,xcdraw]{xcolor} 
\usepackage{pifont}

\newtheorem{finding}{Finding}

\begin{document}

\title{\textsc{KVDrive}: A Holistic Multi-Tier KV Cache Management System for Long-Context LLM Inference}

\author{Jian Lin}
\email{jlindc@connect.ust.hk}
\orcid{0000-0001-9130-7320}
\affiliation{%
  \institution{Hong Kong University of Science and Technology}
  \country{China}
}

\author{Jiazhi Mi}
\email{jmiad@connect.ust.hk}
\orcid{0009-0004-8260-6723}
\affiliation{%
  \institution{Hong Kong University of Science and Technology}
  \country{China}
}

\author{Zicong Hong}
\authornote{Corresponding author}
\email{congcong@ust.hk}
\orcid{0000-0001-5689-382X}
\affiliation{%
  \institution{Hong Kong University of Science and Technology}
  \country{China}
}

\author{Haodong Wang}
\email{hwanghb@connect.ust.hk}
\orcid{0009-0001-8977-850X}
\affiliation{%
  \institution{Hong Kong University of Science and Technology}
  \country{China}
}

\author{Qianli Liu}
\email{qliucc@connect.ust.hk}
\orcid{0009-0007-1292-973X}
\affiliation{%
  \institution{Hong Kong University of Science and Technology}
  \country{China}
}

\author{Haoyue Zhang}
\email{hzhangex@connect.ust.hk}
\orcid{0009-0002-8308-3827}
\affiliation{%
  \institution{Hong Kong University of Science and Technology}
  \country{China}
}

\author{Peng Li}
\email{pengli@xjtu.edu.cn}
\orcid{0000-0002-5303-0700}
\affiliation{%
  \institution{Xi'an Jiaotong University}
  \country{China}
}

\author{Song Guo}
\authornotemark[1]
\email{songguo@cse.ust.hk}
\orcid{0000-0001-9831-2202}
\affiliation{%
  \institution{Hong Kong University of Science and Technology}
  \country{China}
}
% \affiliation{%

% \affiliation{%

% \affiliation{%

% \affiliation{%

% \affiliation{%

% \affiliation{%

\renewcommand{\shortauthors}{Jian Lin et al.}
%%
%% Article type: Research, Review, Discussion, Invited or position
\acmArticleType{Review}
%%
%% Links to code and data
\acmCodeLink{https://github.com/borisveytsman/acmart}
\acmDataLink{htps://zenodo.org/link}
%%
%% Authors' contribution
\acmContributions{BT and GKMT designed the study; LT, VB, and AP
  conducted the experiments, BR, HC, CP and JS analyzed the results,
  JPK developed analytical predictions, all authors participated in
  writing the manuscript.}
%%
%% Sometimes the addresses are too long to fit on the page.  In this
%% case uncomment the lines below and fill them accodingly.
%%
%% \authorsaddresses{Corresponding author: Ben Trovato,
%% \href{mailto:trovato@corporation.com}{trovato@corporation.com};
%% Institute for Clarity in Documentation, P.O. Box 1212, Dublin,
%% Ohio, USA, 43017-6221}
\begin{abstract}
Supporting long-context LLMs is challenging due to the substantial memory demands of the key–value (KV) cache.  
Existing offloading systems store the full cache in host memory and selectively fetch critical entries during decoding, but this strategy quickly hits a ceiling: sparsity cannot be pushed further without degrading accuracy.  
As a result, when context length and batch size grow, the volume of KV transfers rises sharply and becomes the dominant source of decoding latency.  
We present \textsc{KVDrive}, a holistic multi-tier KV cache management system spanning GPU memory, host DRAM, and SSD.  
Unlike prior work that pursues greater sparsity through algorithmic refinements, \textsc{KVDrive} tackles the problem from a systems perspective—jointly orchestrating cache placement, pipeline scheduling, and cross-tier coordination to sustain high-throughput inference under tight GPU budgets.  
\textsc{KVDrive} advances three fundamental capabilities:  
it \emph{adapts cache management to attention behavior} to maximize reuse and minimize redundant data movement;  
it \emph{restructures the decoding pipeline} to overlap I/O- and CPU/GPU compute-bound stages, eliminating stalls across heterogeneous resources;  
and it \emph{harmonizes data movement across memory tiers} to unlock scalable long-context inference far beyond GPU and DRAM limits.  
We have implemented a fully functional prototype of \textsc{KVDrive} and evaluated it on long-context benchmarks with popular LLMs.  
The system achieves up to $1.74\times$ higher throughput compared to state-of-the-art works while preserving accuracy.
\end{abstract}

\begin{CCSXML}
<ccs2012>
   <concept>
       <concept_id>10002951.10002952</concept_id>
       <concept_desc>Information systems~Data management systems</concept_desc>
       <concept_significance>500</concept_significance>
       </concept>
 </ccs2012>
\end{CCSXML}

\ccsdesc[500]{Information systems~Data management systems}
\keywords{Large language model; Long-context serving; KV cache offloading}

\maketitle

\section{Introduction}

As large language models (LLMs) continue to scale in both capability and adoption, supporting long-context inference~\cite{liu2025comprehensivesurveylongcontext} has become increasingly important for applications such as document understanding,  complex agent workflows, software development~\cite{copilot}, and reasoning over large knowledge bases.  

A central obstacle to long-context inference is the memory footprint of the key-value (KV) cache.  
During autoregressive decoding, LLMs must retain the keys and values of all preceding tokens to enable attention computation.  
Unlike model weights, which remain fixed, the KV cache grows linearly with sequence length and batch size, and can easily surpass the size of the model itself.  
Efficient KV cache management has therefore emerged as a critical challenge for enabling long-context LLMs.

% \begin{figure}[t]
%     \centering
%     \includegraphics[width=0.65\linewidth]{figures/workflow1.pdf}
%     \caption{Overview of the computation workflow and storage hierarchy in a KV cache offloading system, showing the key processing stages and the data and resource organization across GPU, DRAM, and SSD.}
%     \label{fig:intro}
% \end{figure}

For example, Llama-3.1-8B-Instruct~\cite{meta_llama_3.1_8b_instruct} supports sequences of up to 128k tokens—roughly 200 pages of text—requiring a KV cache exceeding 16~GB.  
Emerging models extend context windows to 1M tokens~\cite{yang2025qwen251mtechnicalreport}, further amplifying this demand.  
In contrast, commodity GPUs provide only tens of gigabytes of memory, which must also accommodate model weights, activations, and other runtime overheads.  
This memory gap makes it infeasible to store the entire KV cache in GPU memory for long-context or large-batch inference.

To bridge this¥ gap, recent studies propose offloading KV caches to host memory~\cite{tang2406quest,sun2024shadowkv,chen2025magicpig,pqcache,InfiniGen,retroinfer}.
Before each attention computation, the needed entries are reloaded into GPU memory.  
While offloading alleviates GPU pressure, it introduces new inefficiencies across the system.  
Existing approaches fail to coordinate CPU and GPU computation, data transfer, and storage effectively, often leaving one or more resources underutilized—GPU memory idles while waiting for data, CPUs stall during GPU execution, or I/O bandwidth remains unused between transfers.  
As a result, they fail to achieve a holistic balance among GPU, CPU, and I/O subsystems, preventing them from reaching an optimal coordination point.

To address these challenges, we propose \textsc{KVDrive}, a holistic multi-tier KV cache management system spanning GPU memory, host DRAM, and SSD.  
Unlike prior work that relies on algorithmic sparsity refinements, \textsc{KVDrive} approaches the problem from a systems perspective—jointly optimizing cache management, pipeline scheduling, and storage tiering for efficient long-context inference under tight GPU budgets.  
It introduces three key techniques:

\textbf{(1) Attention-Based Cache Management.}  
\textsc{KVDrive} rethinks conventional cache management—traditionally guided by generic access frequency or recency—by making it attention-aware and tailored to the transformer architecture.  
Particularly, although the exact set of critical KV entries varies across tokens, they exhibit temporal locality within decoding windows.  
Leveraging this property, \textsc{KVDrive} maintains a sliding window of critical entries in GPU memory, incrementally updating only out-of-window differences.  
It depends on a 2D layer–head cache allocation and lookahead eviction to maximize reuse while keeping memory overhead bounded.

\textbf{(2) Elastic Pipeline Scheduling.}  
\textsc{KVDrive} decouples selection, fetching, and computation into independently scheduled stages through a new \emph{SFC disaggregation} design.  
Leveraging the distinct characteristics of each stage, the system partitions decoding into micro-batches and executes these stages in parallel with minimal interference.  
Fine-grained micro-batching overlaps the memory-, transfer-, and compute-bound stages, while the index size, cache size, and micro-batch size are jointly tuned to balance latency, throughput, and accuracy.  
This design eliminates pipeline stalls and sustains high utilization across GPU, CPU, and I/O subsystems under diverse workloads.

\textbf{(3) Coordinated Multi-Tier KV Storage.}  
Extending beyond DRAM-only offloading, \textsc{KVDrive} incorporates SSD as a third tier and coordinates data movement across HBM, DRAM, and SSD.  
It applies \emph{importance-guided warm-up} to prioritize high-value KV entries during prefill, employs an \emph{SSD-aware layout} to maximize sequential I/O locality, and performs \emph{parallel sparse synchronization} to minimize cross-tier transfer overhead.  
Together, these mechanisms enable scalable long-context inference well beyond the memory capacity of GPU and DRAM alone.

In summary, this paper makes the following contributions:

\begin{itemize}
\item We present \textsc{KVDrive}, a holistic multi-tier KV cache management system that sustains efficient long-context LLM inference under tight GPU cache budgets.  
\item We introduce an attention-aware cache management mechanism that enables efficient reuse of KV entries in GPU memory, substantially reducing redundant data movement.  
\item We propose an elastic pipeline scheduling strategy that decouples selection, fetching, and computation, achieving fine-grained overlap and eliminating stalls.  
\item We design a coordinated multi-tier storage architecture that provides low-latency access and scalable support for long contexts beyond GPU and DRAM capacity.  
\item We implement and evaluate \textsc{KVDrive} on long-context benchmarks, demonstrating up to $1.74\times$ higher throughput compared to existing systems, while preserving accuracy.
\end{itemize}

\section{Background \& Related Work}

\subsection{KV Cache Basics}

Modern LLMs~\cite{gpt4,llama} are typically built on decoder-only Transformers~\cite{TurboTransformers}.  
Inference is divided into two phases: \textit{prefill} and \textit{decoding}.  
During prefill, all input tokens (the prompt) are processed in parallel. This phase generates the first output token while storing intermediate key and value vectors in GPU memory, collectively referred to as the KV cache.  
In the subsequent decoding phase, tokens are generated autoregressively, with each step appending new key and value vectors to the cache.  
This design eliminates redundant computation and enables efficient token generation.

\subsection{Sparse Attention}

\begin{figure}[t]
    \centering
    \begin{minipage}{0.406\linewidth}
        \centering
        \includegraphics[width=0.95\linewidth]{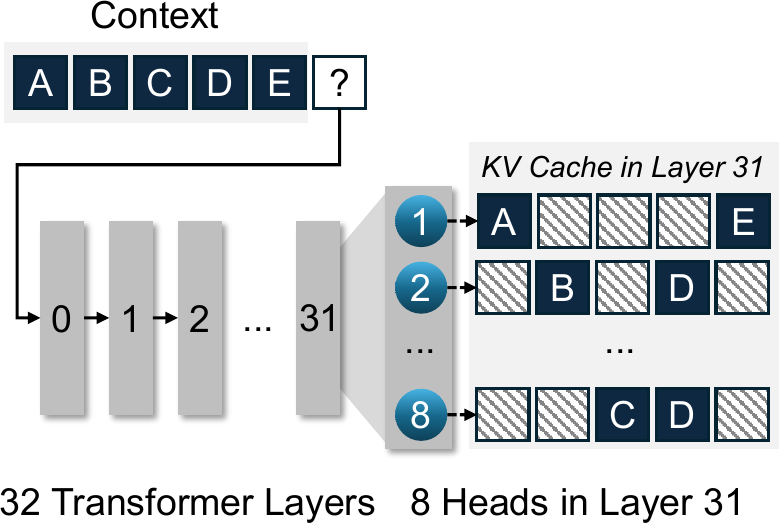}
        \caption{An example of sparse attention in a 32-layer, 8-head model.}
        \label{fig:dsa}
    \end{minipage}
    \hspace{0.05\linewidth}  % 固定间距，替换 \hfill
    \begin{minipage}{0.248\linewidth}
        \centering
        \subfloat[][]{
            \begin{minipage}[t]{\linewidth}
                \centering
                \includegraphics[width=0.855\linewidth]{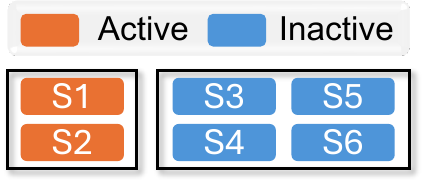}
                \label{fig:existing_work1}
            \end{minipage}
        }\\
        \vspace{-3pt}
        \subfloat[][]{
            \begin{minipage}[t]{\linewidth}
                \centering
                \includegraphics[width=0.855\linewidth]{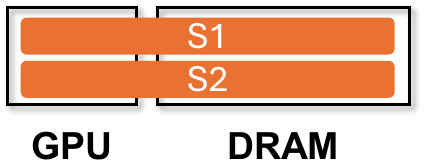}
                \label{fig:existing_work2}
            \end{minipage}
        }
        \caption{Two types of offloading systems.}
    \end{minipage}
\end{figure}

Not all tokens contribute equally to attention: tokens with higher attention scores typically play a more critical role~\cite{flexprefill, tang2406quest, minference,yang2025lserve,deepseekai2024deepseekv32,lu2025mobamixtureblockattention,nsa}.  
Their corresponding key–value pairs are referred to as \textit{critical} KV entries, and discarding non-critical entries generally leads to only minor accuracy loss~\cite{zhang2023h2o}.  
Since token criticality depends on the current query, the system must dynamically identify which KV entries are relevant.  
For example, as shown in \autoref{fig:dsa}, given a context ``ABCDE'', when decoding the next token, Head 1 in Layer 31 identifies the KV entries of tokens ``A'' and ``E'' as critical.  
As a representative system, Quest~\cite{tang2406quest} partitions key entries into chunks and estimates their importance by multiplying the query vector with the channel-wise minimum and maximum of the keys.  
By retrieving only the Top-K important chunks per query, Quest substantially reduces attention computation.

\subsection{KV Cache Offloading}
\label{sec:existing_kv_cache_offloading}

\begin{table*}[t]
\caption{Comparison between several representative KV cache offloading systems and our \textsc{KVDrive}.}
\resizebox{\textwidth}{!}{%
\begin{tabular}{|c|c|c|c|c|c|c|c|}
\hline
                      & Naive          & Quest~\cite{tang2406quest} & ShadowKV~\cite{sun2024shadowkv} & InfiniGen~\cite{InfiniGen} & MagicPIG~\cite{chen2025magicpig} & PQCache~\cite{pqcache} & \textsc{KVDrive} (Ours)           \\ \hline
Caching          & ×              & ×                                                               & LRU & × & ×   & LFU/LRU   & \makecell[c]{\textbf{Attention-Based}\\ \textbf{Cache Management}} \\ \hline
Scheduling & Sequential    & Sequential                              & Sequential                  &  \makecell[c]{Partial Pipeline\\ w/o Fetching Stalls}       & Sequential  & Sequential     & \makecell[c]{\textbf{Elastic Pipeline}\\ \textbf{w/ Minimized Stalls}}             \\ \hline
Tiering & DRAM    & DRAM                              & DRAM                  &  DRAM       & DRAM  & DRAM     & \makecell[c]{\textbf{Coordinated Multi-Tier}\\ \textbf{KV Storage}}            \\ \hline
\end{tabular}%
}
\label{tab:comparison}
\end{table*}

The KV cache size grows linearly with context length and batch size, quickly exceeding GPU memory capacity.  
To mitigate this bottleneck, recent studies propose offloading KV caches from GPU memory to slower but larger tiers such as host memory or SSDs.

Several systems~\cite{gao2024cost, jeong2025accelerating, impress,strata} focus on multi-session scenarios, where each session’s KV cache fits in GPU memory but the aggregate demand across sessions exceeds capacity.  
Inactive-session caches are offloaded while the active session remains on GPU.  
For example, as shown in \autoref{fig:existing_work1}, the KV caches of four inactive sessions (S3–S6) are stored in DRAM while those of two active sessions (S1–S2) remain in GPU memory.  
These methods primarily aim to reduce time-to-first-token.  
However, they assume the active session’s cache always fits in GPU memory—an assumption that fails in long-context inference.

More recent works~\cite{tang2406quest,sun2024shadowkv,chen2025magicpig,pqcache,InfiniGen,retroinfer} instead offload the active sessions' KV caches to host memory, such as S1 and S2 in \autoref{fig:existing_work2}.
During decoding, each token typically requires three steps:  
(1)~\emph{selecting} critical KV entries via the index stored in the GPU memory;  
(2)~\emph{fetching} these entries from host memory to GPU memory;  
(3)~\emph{computing} the new token with sparse attention and other layer operations (e.g., feed-forward networks).

The naive solution is to store all keys in GPU memory as the index and offload all values to host memory.  
Critical entries are then selected by multiplying each query with the full set of keys and retrieving the Top-K.  
To improve efficiency, three optimization strategies have been explored:  
(i)~\textit{Column selection}: InfiniGen~\cite{InfiniGen} selects a subset of key columns with the largest magnitudes.  
(ii)~\textit{Spatial chunking}: Quest~\cite{tang2406quest} and ShadowKV~\cite{sun2024shadowkv} partition adjacent keys into chunks and use their min/max/mean values as representatives.  
(iii)~\textit{Similarity grouping}: MagicPig~\cite{chen2025magicpig} employs Locality-Sensitive Hashing to group similar keys, while RetrievalAttention~\cite{liu2024retrievalattention}, RetroInfer~\cite{retroinfer}, and PQCache~\cite{pqcache} leverage general Approximate Nearest Neighbor Search (ANNS) techniques, which use cluster centroids as index representatives for efficient key retrieval.

As summarized in \autoref{tab:comparison}, \textsc{KVDrive} advances KV cache offloading along three dimensions:
(i)~\textbf{Caching.} Existing systems such as ShadowKV~\cite{sun2024shadowkv}, PQCache~\cite{pqcache}, and RetroInfer~\cite{retroinfer} employ generic cache management policies—typically LFU or LRU—that rely solely on past access frequency or recency.
In contrast, \textsc{KVDrive} introduces an \emph{attention-aware} cache manager that maintains its in-GPU cache based on real-time attention distributions and model-specific architectural patterns, enabling effective reuse aligned with the model’s behavior.
(ii)~\textbf{Scheduling.} Most prior systems follow a sequential pipeline of selection, fetching, and computation, which leads to frequent GPU stalls.
{InfiniGen~\cite{InfiniGen} attempts to mitigate this via speculative prefetching based on the previous layer's attention. However, this approximation often degrades accuracy in long-context tasks.}
\textsc{KVDrive} instead employs an \emph{elastic pipeline scheduler} that overlaps these stages at fine granularity, effectively reducing stalls and sustaining high utilization under diverse workloads without compromising generation quality.
(iii)~\textbf{Tiering.} Prior systems like FlexGen~\cite{flexgen} perform coarse-grained, layer-wise offloading. This is inefficient for sparse attention workloads, as fetching unused KV blocks causes severe I/O amplification. \textsc{KVDrive} instead adopts \emph{parallel sparse synchronization}, fetching only the specific KV blocks required by each query. This fine-grained approach, coupled with our coordinated multi-tier design, allows \textsc{KVDrive} to extend the storage hierarchy to SSDs efficiently, mitigating the bandwidth bottleneck that limits coarse-grained offloading solutions.

These three mechanisms are not isolated optimizations but are integrated to fundamentally change how long-context KV cache management is performed: the elastic scheduler hides the latency of the sparse I/O, while the attention-guided cache minimizes the volume of that I/O, enabling scalable inference cross heterogeneous storage hierarchies.

\section{Motivation}
\label{sec:motivation}

The memory footprint of the KV cache grows linearly with both context length and batch size, which quickly becomes prohibitive in practice.  This linear growth makes directly storing all KV caches in GPU memory impractical under current hardware constraints. For instance, with a batch size of 8 and a context length of 100K, the KV cache of Llama-3-8B requires nearly 100GB of memory—beyond the capacity of most commodity GPUs.

% \begin{figure}[t]
% 	\centering
% 	\subfloat[][Llama-3-8B]{
% 		\begin{minipage}[t]{0.45\linewidth}
% 			\centering
% 			\includegraphics[width=\linewidth]{figures/llama_memory_cost.pdf}
% 		\end{minipage}
% 	}\hfill
%     \subfloat[][Qwen3-14B]{
% 		\begin{minipage}[t]{0.45\linewidth}
% 		\centering
% 		\includegraphics[width=\linewidth]{figures/qwen3_memory_cost.pdf}
% 		\end{minipage}
% 	}
%     \caption{Memory cost of KV caches across different context lengths and batch sizes in two representative models.}
%     \label{fig:llama_mem_cost}
% \end{figure}

As discussed in \autoref{sec:existing_kv_cache_offloading}, prior studies have proposed various KV cache offloading systems.  
We select Quest~\cite{tang2406quest}, RetroInfer~\cite{retroinfer}, and ShadowKV~\cite{sun2024shadowkv} as representative systems, RULER~\cite{hsieh2024ruler} as a benchmark, and an L20 server (\autoref{sec:setup}). 
All baselines are implemented following their original papers; however, to ensure a fair comparison, we disable several auxiliary optimizations.  
Details of these adjustments and the hyperparameters are provided in \autoref{sec:setup}.  
In the following, we identify three fundamental limitations that hinder their practicality and performance in large-scale deployments under realistic long-context and multi-batch settings.

First, most existing systems~\cite{InfiniGen} load a fresh set of critical KV entries from host memory at every decoding step, discarding those previously fetched into GPU.  
Recent studies~\cite{sun2024shadowkv,pqcache} have observed short-range temporal locality—i.e., consecutive queries often attend to overlapping critical KV entries—yet this has not been systematically analyzed in the long range.

% \begin{framed}
% \begin{finding}
% Critical KV entries exhibit strong temporal correlation not only between adjacent tokens but also across a broader local range.  
% Maintaining a sliding window of recent critical KV entries therefore enables effective reuse.
% \end{finding}
% \end{framed}
\begin{center}
\fbox{\parbox{0.95\columnwidth}{%
\begin{finding}
Critical KV entries exhibit strong temporal correlation not only between adjacent tokens but also across a broader local range.  
Maintaining a sliding window of recent critical KV entries therefore enables effective reuse.
\end{finding}
}}
\end{center}

\begin{figure}[t]
	\centering
    \includegraphics[width=0.7\linewidth]{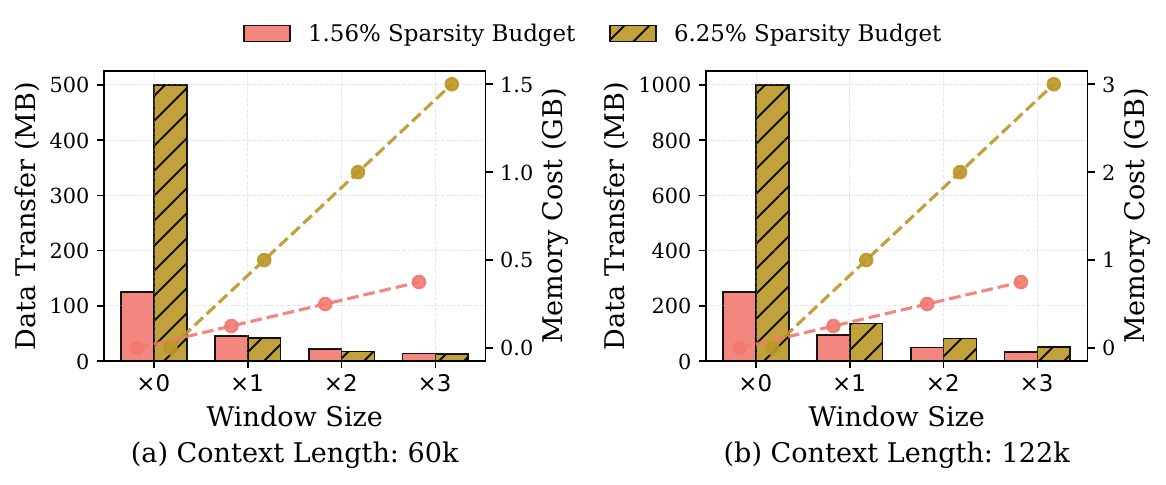}
    \caption{Effect of critical KV windows with different window sizes for Llama-3-8B under $1.56\%$ and $6.25\%$ budgets. Expanding the critical KV window significantly reduces data transfer with minimal memory overhead.}
	\label{fig:Pre_sliding_window_Comm}
\end{figure}

We define a \emph{critical KV window} for each decoding step as the set of critical KV entries corresponding to multiple recent tokens.  
To evaluate temporal reuse, we measure (i) the additional GPU memory overhead and (ii) the amount of reloaded data, defined as the non-overlapping KV entries between adjacent windows.

We denote the window configuration as ``$\times N$'', where $N$ represents how many times the window size exceeds the per-step sparsity budget. ``$\times 0$'' indicates that no KV entries are retained after each decoding step (i.e., the cache is cleared entirely),  
``$\times 1$'' retains one step’s worth of critical entries, and larger values such as ``$\times 2$'' correspond to sliding windows that cover multiple recent steps.

Most existing offloading systems follow a per-step sparsity paradigm:  
each decoding step retrieves its corresponding critical KV entries, uses them once, and then discards them (``$\times 0$'').  
A few methods, such as ShadowKV~\cite{sun2024shadowkv} and PQCache~\cite{pqcache}, retain only the most recent step’s entries (``$\times 1$''), achieving modest reuse. 
As illustrated in \autoref{fig:Pre_sliding_window_Comm}, our analysis shows that maintaining a sliding window of multiple steps (e.g., ``$\times 2$'' and ``$\times 3$'') yields substantially higher benefits.  
With a sparsity budget of $6.25\%$, increasing the window size from $\times 0$ to $\times 3$ reduces host–GPU transfers from over 500~MB to below 12.5~MB per step.  
This observation indicates that a small fraction of additional GPU memory, typically available after accounting for model parameters and intermediate activations,  
can be effectively repurposed to cache recently used KV entries.  

\begin{figure}[t]
    \centering
    \includegraphics[width=0.7\linewidth]{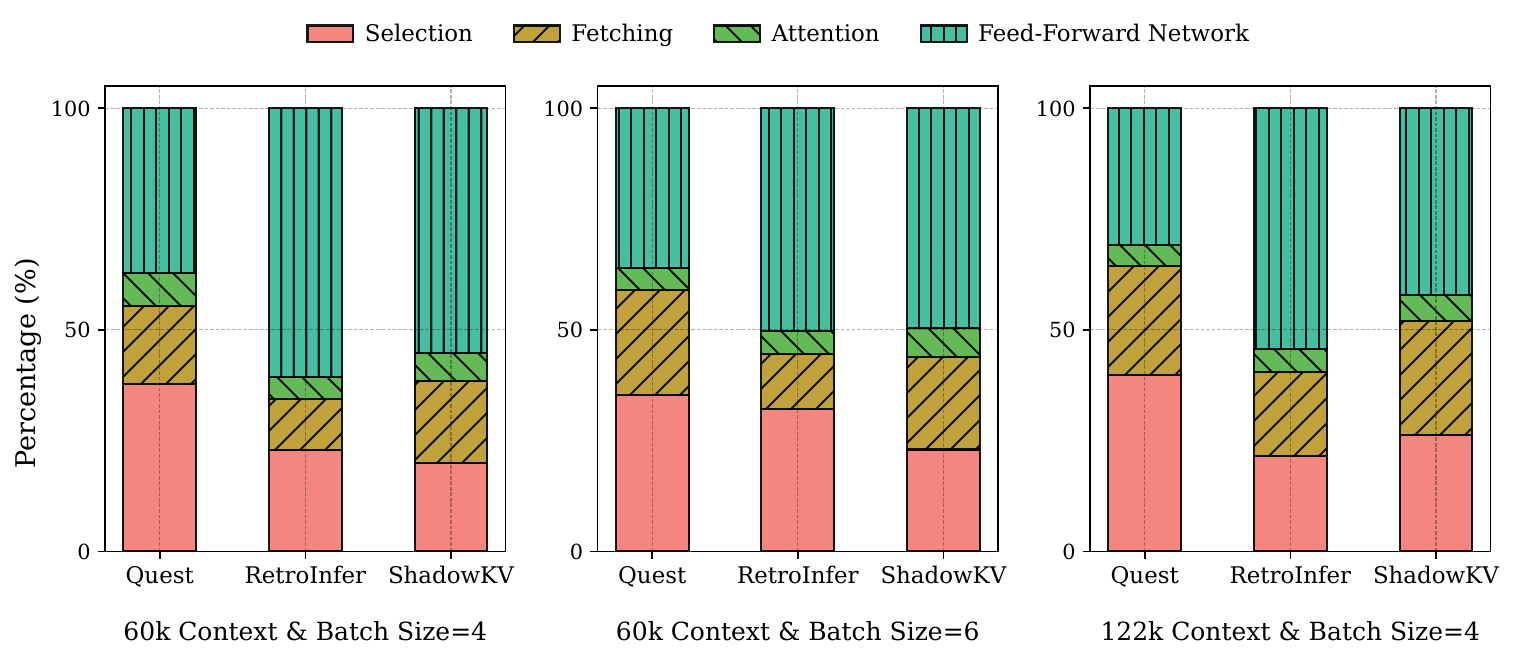}
    \caption{Time breakdown of the three representative offloading systems under different context lengths and batch sizes under a sparsity budget of 1.56\%. KV selection and fetching dominate decoding latency, creating significant GPU stalls that worsen with larger batches and contexts.}
    \label{fig:latency_bottleneck}
\end{figure}

% \begin{framed}
% \begin{finding}
% KV selection and fetching together account for a substantial portion of the decoding latency.
% Their sequential execution creates pronounced GPU pipeline stalls, and their impact increases with larger batch sizes and longer contexts.
% \end{finding}
% \end{framed}

\begin{center}
\fbox{\parbox{0.95\columnwidth}{%
\begin{finding}
KV selection and fetching together account for a substantial portion of the decoding latency.
Their sequential execution creates pronounced GPU pipeline stalls, and their impact increases with larger batch sizes and longer contexts.
\end{finding}
}}
\end{center}

As shown in \autoref{fig:latency_bottleneck}, both KV selection and fetching together account for nearly 50\% of the total runtime under realistic context lengths and batch sizes.  
This bottleneck arises mainly from the high computational cost of selection and the data movement in fetching, and it becomes increasingly pronounced as the batch size or context length grows.  
During selection, different systems adopt distinct indexing schemes with varying computational overhead.  
Quest~\cite{tang2406quest} evaluates each query against both the minimum and maximum keys within every chunk, resulting in the longest selection latency.  
In contrast, ShadowKV~\cite{sun2024shadowkv} compares queries only with the mean key of each chunk, while RetroInfer~\cite{retroinfer} uses cluster centroids, both of which reduce the selection cost.  
Following selection, the fetching stage must transfer the identified KV entries from host memory to GPU memory before attention computation can proceed.  
Because most existing systems execute these stages sequentially—select, then fetch, then compute—both phases block GPU execution and cause substantial stalls.  
Consequently, although offloading alleviates GPU memory pressure, these stalls dominate overall runtime at scale and prevent existing systems from efficiently handling long contexts or large batches.

% \begin{framed}
% \begin{finding}
% Due to the limited bandwidth between GPU and disk compared to GPU–DRAM transfers, existing offloading systems struggle to efficiently migrate KV caches to disk, leading to severe throughput degradation.
% \end{finding}
% \end{framed}

\begin{center}
\fbox{\parbox{0.95\columnwidth}{%
\begin{finding}
Due to the limited bandwidth between GPU and disk compared to GPU–DRAM transfers, existing offloading systems struggle to efficiently migrate KV caches to disk, leading to severe throughput degradation.
\end{finding}
}}
\end{center}

\begin{figure}[t]
	\centering
	\subfloat[][]{
		\begin{minipage}[t]{0.343\linewidth}
			\centering
			\includegraphics[width=\linewidth]{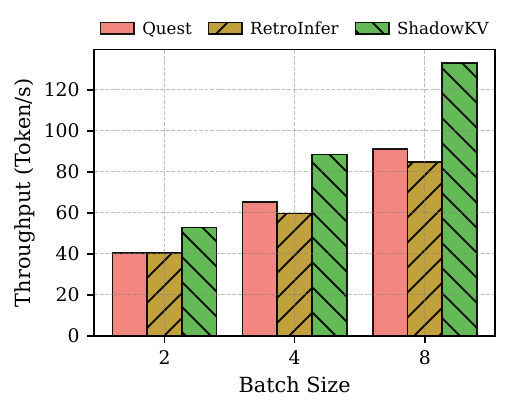}
		\end{minipage}
        \label{fig:throughput_scaling_1}
	}
        \subfloat[][]{
		\begin{minipage}[t]{0.343\linewidth}
		\centering
		\includegraphics[width=\linewidth]{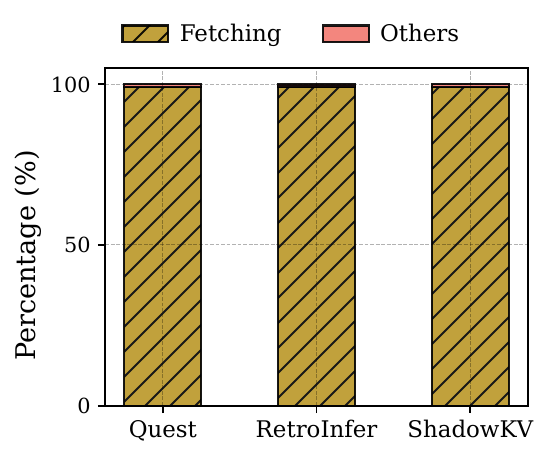}
		\end{minipage}
        \label{fig:throughput_scaling_2}
	}
    \caption{Throughput scaling under DRAM-only and disk-backed offloading for a batch size of 8 and 122k context.  
        (a)~Throughput increases with batch size.  
        (b)~A strawman operates beyond DRAM limits but suffers from severe GPU–SSD bandwidth bottlenecks.}
    \label{fig:throughput_scaling}
\end{figure}

As shown in \autoref{fig:throughput_scaling_1}, when ample host memory is available, increasing the batch size significantly improves throughput, since wider batches better amortize memory accesses and parallelize computation.  
However, under realistic deployment constraints—where a typical data-center GPU node is provisioned with around 100~GB of host memory~\cite{ndasra100,gcp} and edge devices often provide even less (tens of gigabytes)—the KV cache quickly exhausts available capacity once both long contexts and large batches are considered.  
Throughput then plateaus and eventually fails due to out-of-memory errors when host memory becomes saturated.
Thus, systems that offload KV caches only to DRAM cannot fully exploit GPU compute power: memory pressure inevitably forces cache spillover to SSD, and without effective multi-tier management, decoding latency once again becomes dominated by data movement rather than computation.
For example, as illustrated in \autoref{fig:throughput_scaling_2}, we implement a strawman design inspired by FlexGen~\cite{flexgen}, which stores the KV cache on disk using memory mapping.  
During inference, each layer loads its corresponding KV entries from SSD into GPU memory on demand, performs attention computation, and immediately evicts them back to disk after use.  
While this approach enables operation beyond DRAM capacity, it exposes the severe bandwidth gap between GPU–SSD and GPU–DRAM transfers, resulting in extremely limited throughput and frequent GPU stalls.

\section{System Overview}
\label{sec:system_overview}

\begin{figure}[t]
    \centering
    \includegraphics[width=0.7\linewidth]{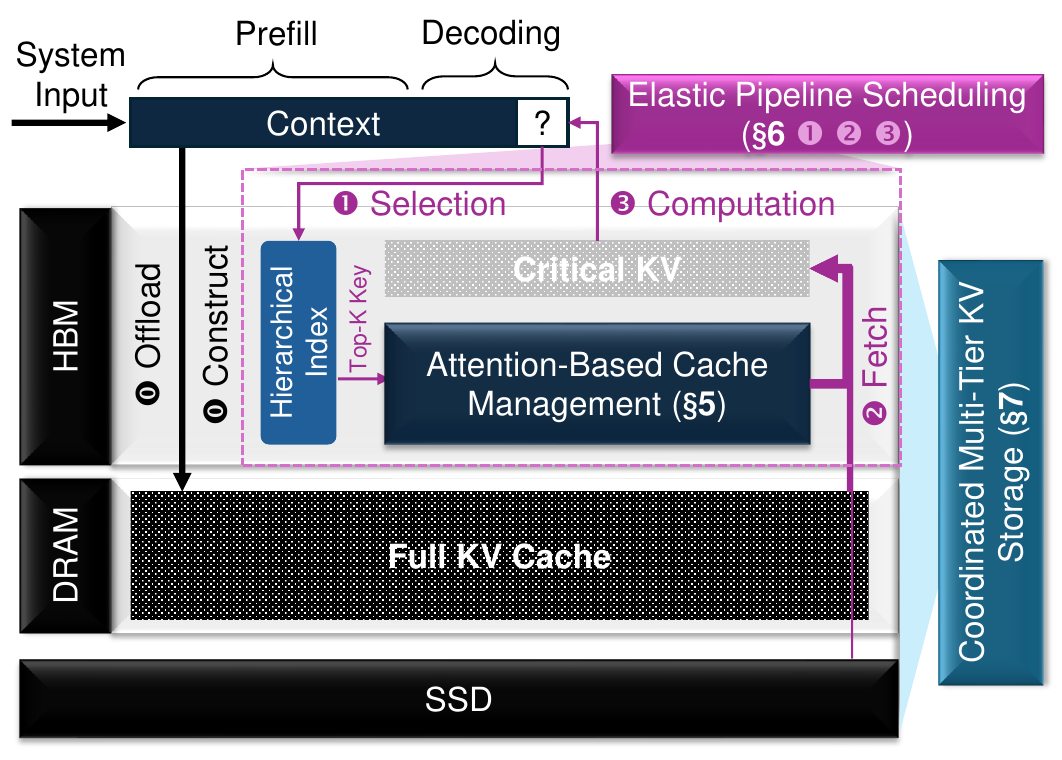}
    \caption{System architecture. During the prefill phase, the system offloads the full KV cache to DRAM/SSD and constructs a hierarchical index. In the decoding phase, the Elastic Pipeline Scheduling module orchestrates a three-stage pipeline (\ding{182}-\ding{184}). When the Attention-Based Cache Management identifies missing critical KV entries, the scheduler fetches the required data from the Coordinated Multi-Tier KV Storage (DRAM/SSD) into HBM. Finally, attention and feed-forward computations are executed using both the newly fetched and resident KV entries.}
    \label{fig:sys_overview}
\end{figure}

As illustrated in \autoref{fig:sys_overview}, \textsc{KVDrive} aims to support high-throughput long-context LLM inference despite tight GPU memory constraints.  
When the KV cache exceeds GPU capacity, it is offloaded to host DRAM or SSDs and the system constructs an index in GPU memory during the prefill phase.  
During decoding, each new token follows a three-stage workflow:  identifying critical KV entries via the index(\ding{182});  fetching the selected entries from DRAM or SSD into GPU HBM (\ding{183}); and executing attention and feed-forward computations over the union of the newly fetched and resident KV entries (\ding{184}). 

\textsc{KVDrive} adopts indexing and sparse attention mechanisms following best practices from prior work (\autoref{sec:existing_kv_cache_offloading})—combining spatial chunking and similarity grouping through a new \emph{hierarchical} design that organizes a lightweight index in a content-aware manner.  
Specifically, the KV cache is partitioned into chunks, and the mean key of each page is used as its representative~\cite{sun2024shadowkv,tang2406quest}, forming higher-level centroids for similarity grouping.  
Unlike global K-means–based ANNS approaches~\cite{liu2024retrievalattention}, this hierarchical structure preserves local semantic continuity among contiguous tokens.  
According to \autoref{sec:experiment}, it achieves higher retrieval accuracy than similarity-grouping methods. Furthermore, compared to spatial chunking, it matches retrieval precision while reducing index footprint by 50\% and accelerating lookup speeds by up to $2\times$ faster lookup speed, effectively reducing GPU memory pressure and facilitating downstream scheduling and tiering optimizations.

\textsc{KVDrive} is structured around three core components:  

\textbf{(1) Attention-Based Cache Management:}  
\textsc{KVDrive} goes beyond naive caching through two complementary mechanisms: 
(i) a \emph{lookahead eviction policy} that leverages current attention signals to anticipate near-future reuse, ensuring that entries most likely to be needed remain resident; and  
(ii) a \emph{2D layer–head scaling strategy} that allocates per-layer and per-head window sizes according to measured attention locality.  
Together, these mechanisms maximize reuse under tight GPU memory budgets while maintaining a fixed overall footprint (\autoref{sec:KV_cache_management}).

\textbf{(2) Elastic Pipeline Scheduling:}  
Redesigns the decoding pipeline through two complementary techniques:  
(i) \emph{SFC disaggregation}, which decouples selection, fetching, and computation into independently scheduled stages, enabling fine-grained overlap between operations; and  
(ii) \emph{pipeline optimization}, which optimizes index size, cache size, and micro-batch size to balance accuracy and throughput.  
Together, these mechanisms eliminate pipeline stalls and maintain high utilization across GPU, CPU, and I/O subsystems under varying batch and context configurations (\autoref{sec:Preftcher}).

\textbf{(3) Coordinated Multi-Tier KV Storage:}  
Extends beyond DRAM-only offloading by incorporating SSD as a third tier and coordinating data movement across HBM, DRAM, and SSD.  
\textsc{KVDrive} (i) applies \emph{importance-guided warm-up} to prioritize high-value KV entries during prefill, (ii) employs an \emph{SSD-aware layout} to maximize sequential I/O locality, and (iii) performs \emph{parallel sparse synchronization} to minimize tier transfer overhead.  
Together, these mechanisms enable scalable long-context inference well beyond the memory capacity of GPU and DRAM alone (\autoref{sec:ssd_tiering}).

\section{Attention-Based Cache Management}
\label{sec:KV_cache_management}

This section presents an in-GPU cache management scheme that departs from traditional usage-based policies (e.g., LRU, LFU) by leveraging the model’s attention mechanism to infer the semantic importance of cached entries.
By aligning cache residency with attention-derived importance rather than mere access recency or frequency, \textsc{KVDrive} captures temporal locality across nearby tokens and supports incremental updates instead of redundant reloads.

\subsection{Sliding Window w/ Lookahead Eviction}
\label{sec:lookahead_eviction}

\begin{figure}[t]
    \centering
    \subfloat[][K=256]{
        \begin{minipage}[t]{0.343\linewidth}
			\centering
        \includegraphics[width=\linewidth]{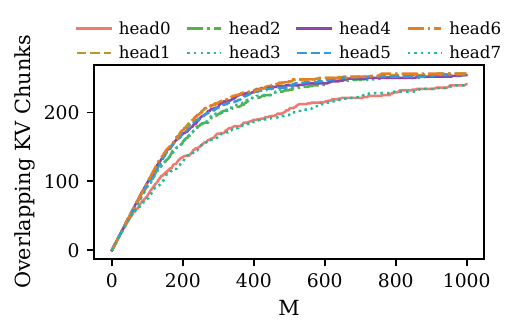}
        \end{minipage}

    }
    \subfloat[][K=512]{
        \begin{minipage}[t]{0.343\linewidth}
			\centering
        \includegraphics[width=\linewidth]{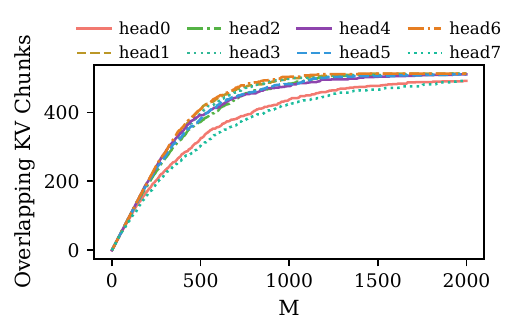}
        \end{minipage}
    }
    \caption{Number of Top-M critical KV entries at one decoding step that also belong to the Top-K set at the next step.}
    \label{fig:topM_overlap}
\end{figure}

\textsc{KVDrive} maintains a sliding window of critical KV entries covering several recent tokens in GPU memory.
At initialization, offline profiling establishes a mapping between the window size and memory footprint for a given model.
The system then selects the largest feasible window size that fits within the available GPU cache budget, where the budget is determined after accounting for the memory occupied by model parameters, activations, and intermediate buffers.
This ensures that the cache fully utilizes the remaining GPU capacity without interfering with model computation.

During decoding, the window advances by one token at a time: new critical KV entries are fetched from host memory, while a subset of existing ones are evicted.
\autoref{fig:topM_overlap} shows that entries receiving high attention at one step are much more likely to remain critical in the next.
As $M$ increases, the overlap with the subsequent step’s top-$K$ set rises rapidly at first and then saturates, indicating that highly ranked entries are consistently reused while the marginal benefit of including lower-ranked ones quickly diminishes.

Motivated by this observation, instead of traditional LRU or LFU policies, \textsc{KVDrive} adopts a \emph{lookahead eviction policy}: at each step, entries with the lowest current-step attention scores are discarded, as they are least likely to be reused in subsequent steps.

By maintaining a compact and incrementally updated working set of KV entries, \textsc{KVDrive} amortizes host–GPU transfers across decoding steps, improving bandwidth efficiency and reducing redundant data movement.

\subsection{2D Window Scaling}
\label{sec:2D_scaling}

% \begin{figure}[t]
% 	\centering
%     \includegraphics[width=0.7\linewidth]{figures/2d_window_scale.pdf}
%     \caption{Data transfer and memory overhead for different window sizes across layers and heads.}
%     \label{fig:layer_wise}
% \end{figure}

% \begin{figure}[t]
%     \centering
%     \includegraphics[width=0.6\linewidth]{figures/window.pdf}
%     \caption{The offline initialization and online running of the adaptive in-GPU cache management in \textsc{KVDrive}.}
%     \label{fig:window}
% \end{figure}

\begin{figure*}[t]
    \centering
    \begin{minipage}{0.45\linewidth}
    \centering
    \includegraphics[width=\linewidth]{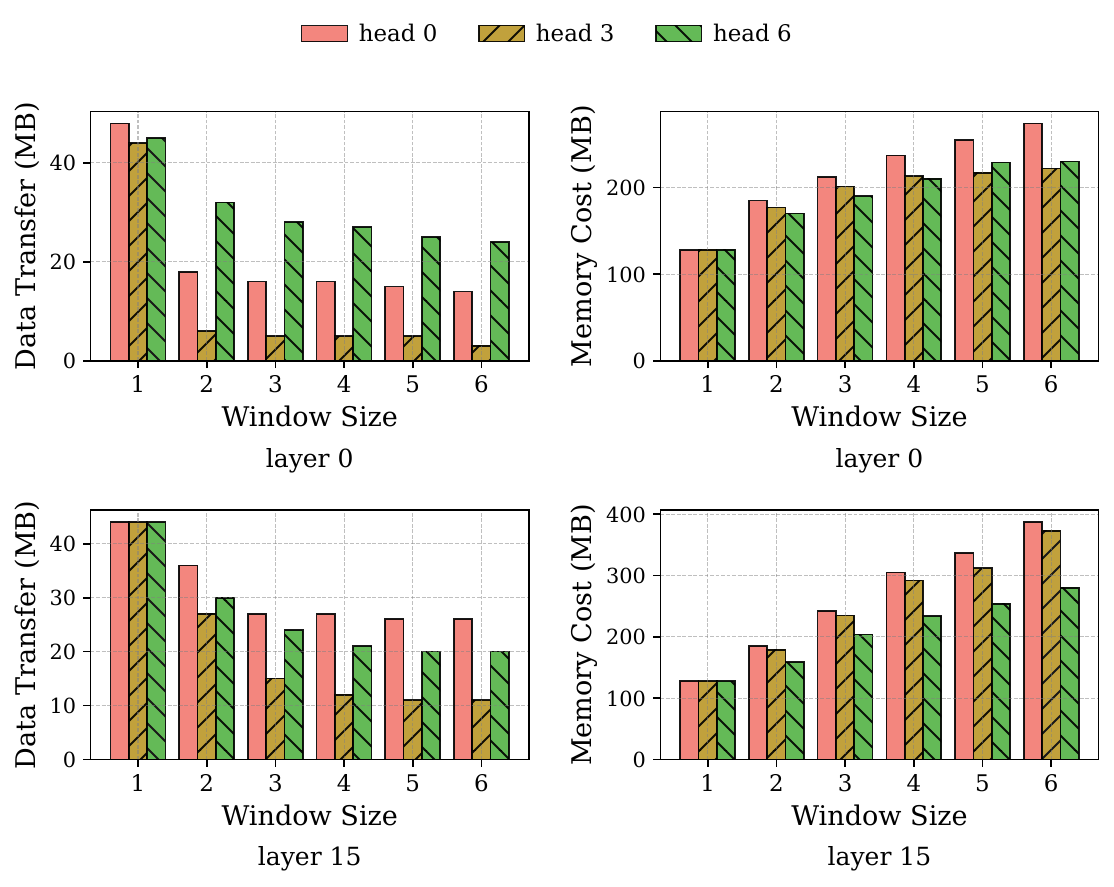}
    \caption{Data transfer and memory overhead for different window sizes across layers and heads.}
    \label{fig:layer_wise}
    \end{minipage}
    \hfill
    \begin{minipage}{0.49\linewidth}
    \centering
    \includegraphics[width=\linewidth]{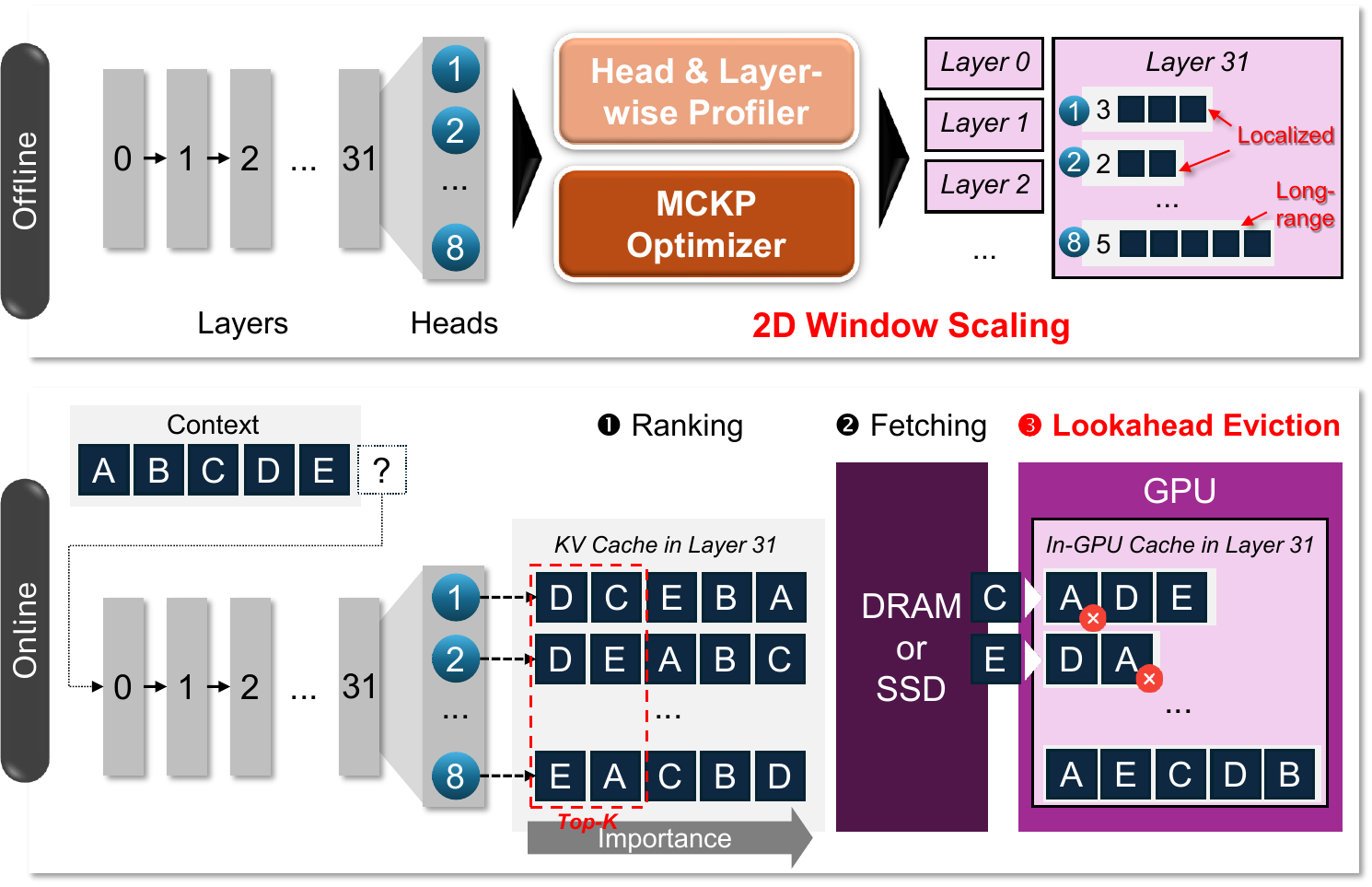}
    \caption{The offline initialization and online running of the adaptive in-GPU cache management in \textsc{KVDrive}.}
    \label{fig:window}
    \end{minipage}
\end{figure*} 

Our profiling in \autoref{fig:layer_wise} shows that different layers and attention heads exhibit heterogeneous trade-offs between memory overhead and transfer reduction when enlarging the window size.  
For most layer–head pairs, larger windows provide only modest savings, whereas certain layers and specific heads achieve disproportionately higher reductions in transfer volume for the same memory cost.  
This reflects the diverse roles of transformer layers and heads: some primarily capture localized dependencies, while others specialize in modeling long-range structure.
We should allocate less space for the former one (e.g., Head~1 and 2 in Layer~31 in \autoref{fig:window}) and more for the latter one (e.g., Head~8 in Layer~31 in \autoref{fig:window}).

To systematically exploit this heterogeneity, we formulate \emph{2D window scaling} as an offline optimization problem.  
For each layer $l$ and head $h$, profiling yields:  
$\text{Benefit}_{l,h}(w)$: transfer reduction achieved with window size $w$,  
$\text{Cost}_{l,h}(w)$: additional GPU memory consumed by that window size. 

Given a total GPU cache budget $M$, the objective is:

\[
\max_{\{w_{l,h}\}} \sum_{l,h} \text{Benefit}_{l,h}(w_{l,h})
\quad \text{s.t.} \quad
\sum_{l,h} \text{Cost}_{l,h}(w_{l,h}) \leq M.
\]

This allocation problem is a variant of the \emph{multiple-choice knapsack problem} (MCKP), which is NP-hard in general.  
Fortunately, the problem size in our setting is moderate (hundreds of layer–head pairs and a small set of candidate window sizes).  
In practice, \textsc{KVDrive} solves it offline: for small models, exhaustive search is feasible; for larger ones, we employ a greedy algorithm that starts from the smallest windows and iteratively enlarges the window with the highest benefit-to-cost ratio until the GPU cache budget is met.  
This approach produces near-optimal allocations within minutes and incurs no runtime overhead.

\begin{figure*}[t]
    \centering
    \begin{minipage}{0.67\linewidth}
    \centering
    \includegraphics[width=\linewidth]{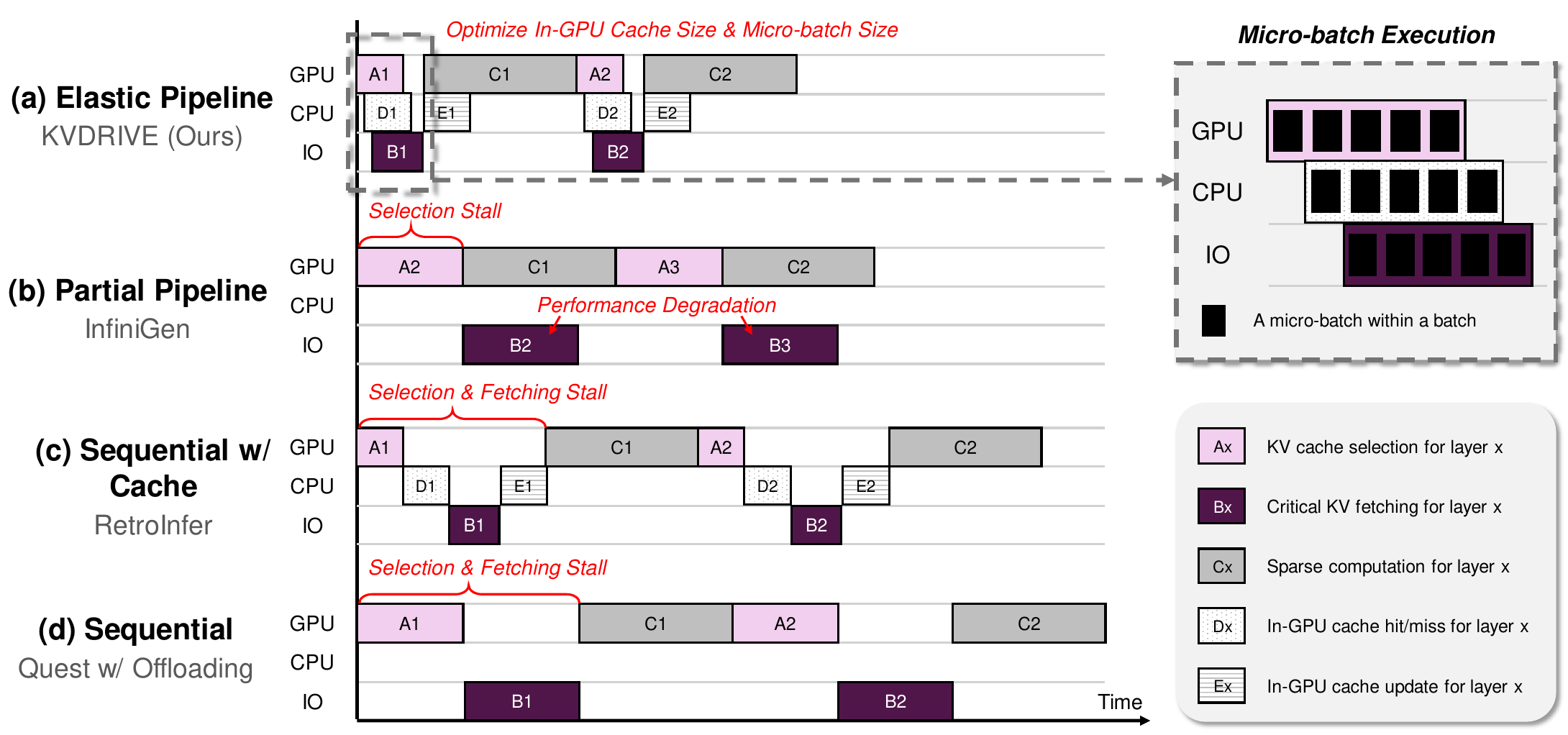}
    \caption{Comparison between different KV cache offloading scheduling strategies.}
    \label{fig:pipeline_scheduling}
    \end{minipage}
    \hfill
    \begin{minipage}{0.3\linewidth}
    \centering
    \includegraphics[width=0.93\linewidth]{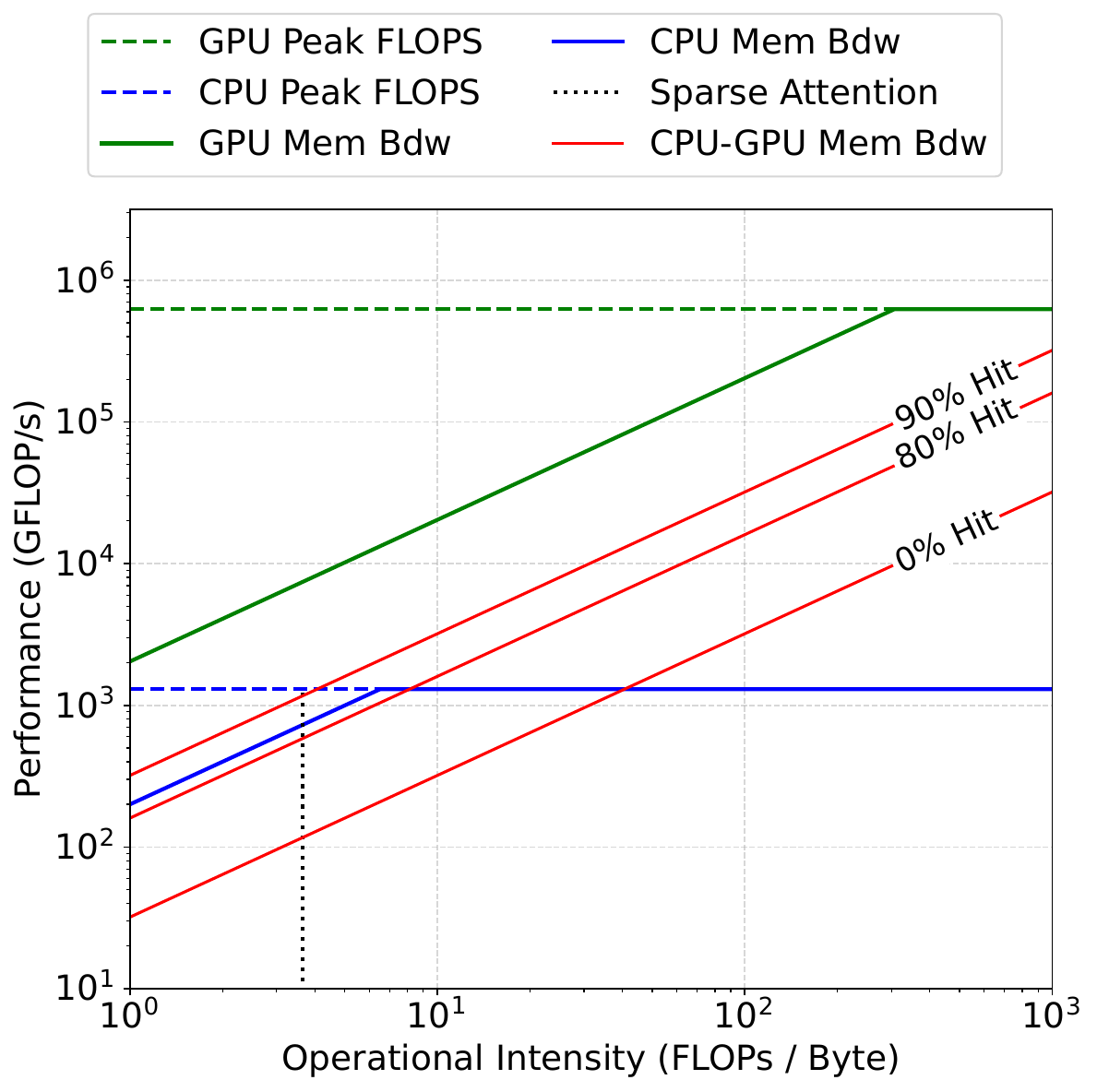}
    \caption{A GPU-CPU roofline model of Llama-3-8B in a KV cache offloading system on an A100 instance.}
    \label{fig:roofline}
    \end{minipage}
\end{figure*} 

By tailoring window sizes across both layers and heads, \textsc{KVDrive} achieves finer-grained cache allocation, striking a better balance between memory efficiency and communication cost, thereby enabling high-quality inference under strict resource constraints.

\section{Elastic Pipeline Scheduling}
\label{sec:Preftcher}

Most existing systems~\cite{tang2406quest,chen2025magicpig,sun2024shadowkv} adopt a sequential workflow (\autoref{fig:pipeline_scheduling}c,d), where the computation of each layer during decoding proceeds in three stages: selecting critical KV entries, fetching them from host memory, and executing the layer’s operations on the GPU. This design introduces substantial \emph{stalls}, i.e., idle GPU cycles when computation must wait for selection or data transfer. To mitigate stalls, InfiniGen~\cite{InfiniGen} adopts a pipelined design in which each layer prefetches critical KV entries using attention input from the previous layer (\autoref{fig:pipeline_scheduling}b). This reduces fetching stalls by overlapping data transfer with computation, but selection stalls remain unresolved, and the approximation used for prefetching can lead to suboptimal KV selection, potentially degrading accuracy and stability in long-context scenarios.  
To eliminate both types of stalls without sacrificing accuracy, we propose an elastic pipeline scheduling strategy, as detailed below.

\subsection{SFC Disaggregation}

The three tightly coupled stages—selection, fetching, and computation—exhibit distinct performance bottlenecks.  
The selection stage executes on the GPU and is primarily I/O-bound, as it requires reading and scoring large index regions.  
The fetching stage is dominated by host–device data transfers, while computation (mainly feed-forward operations) is compute-bound on the GPU.  
Moreover, in-GPU cache hit/miss evaluation and metadata updates described in \autoref{sec:KV_cache_management} require CPU participation, introducing additional synchronization overhead.  
Executing these stages sequentially and applying an identical batching strategy across them leads to suboptimal utilization of GPU and CPU compute units as well as I/O bandwidth.

\textsc{KVDrive} addresses these inefficiencies through \emph{SFC disaggregation}, which decouples the three stages for independent scheduling.  
Each batch is partitioned into multiple micro-batches to enable fine-grained parallelism between selection and fetching, while computation is executed for the entire batch as a single unit.  
During decoding, the GPU performs selection for the current micro-batch concurrently with the CPU evaluating cache hit/miss status for the previous one and fetching the KV entries of an earlier micro-batch from host memory.  
Computation proceeds once all micro-batches have completed fetching.  
Meanwhile, cache metadata updates are overlapped with computation, allowing the system to maintain cache consistency without interfering with the main decoding pipeline.  
This disaggregated design sustains high throughput by maximizing utilization across GPU, CPU, and I/O subsystems.

Unlike conventional pipeline overlap, SFC disaggregation explicitly separates the three stages into independently scheduled units coordinated through lightweight queues and asynchronous transfers.  
This design maintains balanced utilization across heterogeneous resources and delivers high throughput.

\subsection{Pipeline Optimization}

The performance of \textsc{KVDrive}'s pipeline depends critically on three parameters: the number of centroids in the index, the GPU cache size, and the micro-batch size.  
These parameters jointly determine the balance between selection accuracy, CPU–GPU coordination cost, and overall pipeline efficiency.

\textbf{(1) Index.}  
A larger number of centroids improves selection granularity but increases selection cost, as each query must compare against more index representatives.  
Conversely, too few centroids reduce accuracy in identifying critical KV entries.  
Our experiments show that \textsc{KVDrive} achieves comparable accuracy to spatial-chunking methods~\cite{tang2406quest,sun2024shadowkv} using only half as many centroids, significantly reducing selection latency.

\textbf{(2) Cache size.}  
A larger in-GPU cache shortens fetching time by reducing host–device transfers, but increases CPU-side hit/miss evaluation time, which can become the new bottleneck.  
Before decoding begins, \textsc{KVDrive} performs a warm-up phase that incrementally increases the cache size until the CPU evaluation time and GPU fetching time reach equilibrium, maximizing end-to-end throughput under given memory constraints.

\textbf{(3) Micro-batch size.}  
The micro-batch size determines pipeline granularity.  
Excessively large micro-batches create long bubbles between overlapping stages, whereas very small ones lead to frequent kernel launches and synchronization overhead.  
Since the candidate range is limited, \textsc{KVDrive} performs a short pre-run calibration to empirically identify the optimal configuration before inference begins.
Through this lightweight tuning process, the system adapts its scheduling and resource allocation to the underlying hardware characteristics, ensuring balanced utilization across compute and I/O components.

\subsection{Performance Analysis via Roofline Model}
\label{sec:roofline}

Prior works~\cite{moe_lightning,chen2025magicpig} advocate performing attention on the CPU in KV cache offloading systems, arguing that CPU–GPU transfer overhead dominates GPU execution. We instead analyze why GPU-based attention is preferable in our system, using the roofline model~\cite{yuan2024llminferenceunveiledsurvey}.

\autoref{fig:roofline} shows the roofline model for attention computation in Llama-3-8B when all KV caches reside in CPU memory. When the operational intensity falls below the threshold $P$, transferring data to the GPU yields no benefit, as performance is bounded by the CPU–GPU bandwidth roof. This is the regime assumed in prior studies.  
In contrast, our adaptive cache management in \autoref{sec:KV_cache_management} ensures that at each decoding step, only out-of-cache critical KV entries—those not already in the in-GPU cache—must be fetched from host memory. Empirical evaluations demonstrate that even under a strictly constrained cache budget, approximately 80\% of critical entries are served directly from the in-GPU cache (see \autoref{tab:hit_rate} in \autoref{sec:experiment}). This high hit rate minimizes off-chip data movement, ensuring the operational intensity remains above the threshold 
P. Consequently, the GPU maintains a higher effective throughput than the CPU, making it the preferable choice for attention computation.

\begin{figure*}[t]
    \centering
    \includegraphics[width=\linewidth]{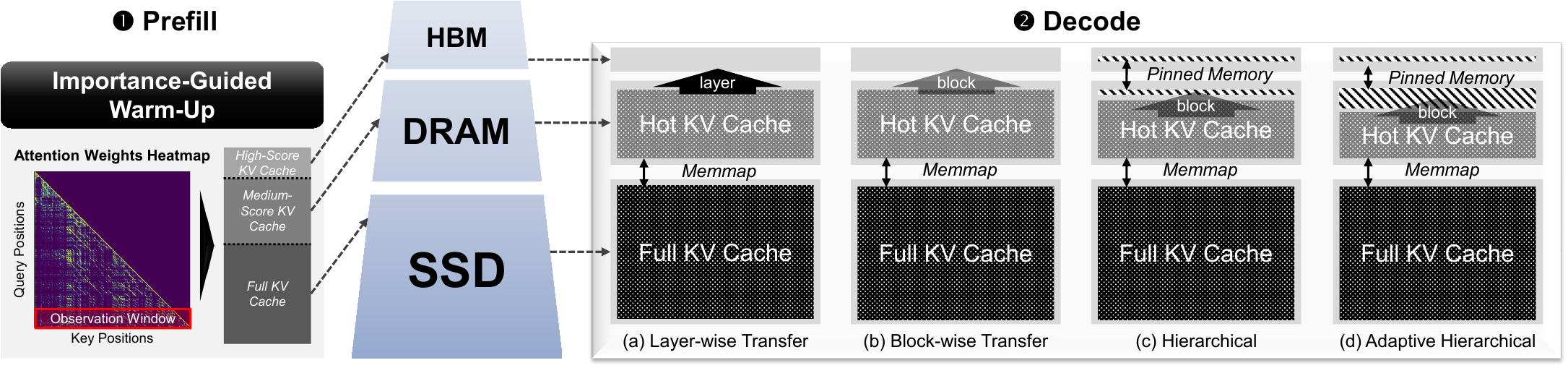}
    \caption{The workflow of the coordinated multi-tier KV storage in \textsc{KVDrive}.}
    \label{fig:tiering}
\end{figure*}

\section{Coordinated Multi-Tier KV Storage}
\label{sec:ssd_tiering}

While GPU and DRAM offer high-bandwidth access for KV caches, their combined capacity quickly becomes insufficient under long contexts or large batches.  
To extend capacity, prior work explores offloading KV caches to SSDs~\cite{io_ssd}; yet directly treating SSD as a slower extension of DRAM causes frequent high-latency transfers and stalls during decoding.  
To address this limitation, \textsc{KVDrive} designs a coordinated multi-tier KV storage system that enables efficient collaboration across HBM, DRAM, and SSD.  
It integrates three complementary techniques: importance-guided warm-up for tier placement initialization, an SSD-aware layout for sequential I/O locality, and a parallel sparse synchronization pipeline for low-latency data movement during decoding.

\subsection{Importance-Guided Warm-Up}

At the end of the prefill phase, the model has already computed the attention between the prompt’s final tokens and the entire prefix.  
\textsc{KVDrive} exploits this information to estimate the long-term importance of prefix KV entries before decoding begins.  
Inspired by SnapKV~\cite{SnapKV}, we assign importance scores to prefix tokens based on the attention distribution produced by the queries in the prompt’s final observation window.

The observation window refers to the last few tokens of the prompt (typically 16–64 tokens).  
For each query within this window, we compute its attention weights over all prefix keys and aggregate them across heads and layers to obtain an importance profile of prefix positions.  
This aggregated profile captures which KV entries are most likely to be reused during decoding and directly guides their initial placement across memory tiers.

According to this importance profile, all KV entries are first persisted to SSD as the full backing store, while higher-ranked entries are promoted to faster tiers:  
the entries with the highest scores are placed in GPU HBM for immediate access; the subsequent highest-scoring entries—those that do not fit in HBM—are offloaded to DRAM. 
This one-time, attention-informed warm-up establishes a balanced starting point for decoding, significantly reducing subsequent data migration across tiers.  
Unlike SnapKV, which leverages observation windows for KV compression, our approach repurposes the same insight for tiered placement, enabling coordinated HBM–DRAM–SSD management without compromising model accuracy.

\subsection{SSD-Aware Layout Planning}

When offloading KV entries to SSD, the data layout becomes a critical determinant of I/O efficiency.  
Random accesses on SSDs are costly, whereas sequential operations can achieve an order of magnitude higher throughput.  
To align storage organization with the temporal and structural access patterns observed during decoding, \textsc{KVDrive} employs a two-level packing strategy.

We first define an \emph{extent} as a contiguous SSD block that groups multiple KV entries together.  
By packing entries into extents, the system can transform fine-grained, irregular accesses into coarse-grained sequential transfers, improving bandwidth utilization and reducing access latency.

\textbf{(1)~Semantic-Contiguity Packing.}  
KV entries that are frequently attended together during decoding—such as tokens within the same semantic chunk or attention cluster—are placed sequentially within the same extent.  
This organization enables multiple related entries to be retrieved through a single large I/O, minimizing random-access overhead across non-contiguous regions.

\textbf{(2)~Layer–Head Partitioning.}  
To further align with Transformer structure, extents are partitioned by layer and attention head.  
Each \{layer, head\} pair is assigned to a dedicated SSD segment, and its extents are stored contiguously within that segment to preserve structural locality.

By combining semantic-contiguity packing with layer–head partitioning, \textsc{KVDrive} reshapes inherently irregular KV access patterns into predictable, high-throughput sequential I/O.  
This SSD-aware layout design substantially reduces access latency and improves end-to-end decoding efficiency, making SSD-based tiering practical even for long-context and large-batch workloads.

\subsection{Parallel Sparse Synchronization}

Efficient data transfer across SSD, DRAM, and GPU HBM is essential for sustaining decoding throughput under multi-tier offloading.  
While SSD offers large capacity, its bandwidth is significantly lower than that of DRAM, and frequent migrations can easily become the new bottleneck.  
\autoref{fig:tiering} (right) compares four synchronization strategies that progressively improve data movement efficiency.

\textbf{(a) Naïve Layer-wise Transfer.}  
This follows FlexGen~\cite{flexgen}: at each decoding step, the KV cache of an entire layer is transferred from SSD to HBM, using DRAM (via \texttt{memmap}) as a passive buffer.  
Although simple, this design provides almost no reuse: when the DRAM cache is smaller than the total KV footprint, subsequent layer transfers overwrite previously cached data, causing redundant I/O and frequent pipeline stalls as discussed in \autoref{sec:motivation}.

\textbf{(b) Block-level Sparse Fetching.}  
Instead of migrating entire layers, \textsc{KVDrive} transfers only the blocks (clusters) of KV entries that are actually required by the current attention queries.  
This fine-grained selection reduces redundant transfers and better matches the sparsity patterns of attention.  
To avoid excessive random I/O, multiple block requests are coalesced into extent-sized sequential reads with sufficient queue depth, ensuring high SSD throughput and efficient utilization of the I/O pipeline.

\textbf{(c) Hierarchical Transfer.}  
To further reduce SSD latency and overlap data preparation with computation, \textsc{KVDrive} adopts a hierarchical staging pipeline.  
KV entries fetched from SSD are first read into the DRAM-backed \texttt{memmap} region and then staged into preallocated page-pinned buffers before GPU execution.  
This design serves two purposes: (i) it avoids repeatedly allocating pinned memory by reusing a stable pool of pinned buffers, and (ii) it enables asynchronous prefetching—KV entries can be read ahead during the selection and pre-attention (QKV projection).  

\textbf{(d) Balanced Coordination.}  
Finally, \textsc{KVDrive} balances the benefits of pinned buffers and \texttt{memmap} caching.  
Pinned memory provides high-bandwidth access, while the \texttt{memmap} region serves as a large passive cache in DRAM.  
The allocation between these two tiers is guided by offline profiling, prioritizing pinned memory for layer-head KV entries exhibiting frequent stalls—patterns empirically correlated with page-fault-heavy areas in the \texttt{memmap} tier—and keeping such regions resident in pinned memory to ensure stable high-bandwidth access throughout decoding.

Overall, this parallel sparse synchronization enables efficient multi-tier coordination across HBM, DRAM, and SSD, sustaining high GPU utilization under long-context and large-batch settings.

\section{Implementation}

We have implemented a fully functional prototype of \textsc{KVDrive} comprising about 9{,}000 lines of Python, 1{,}000 lines of C++, and 3{,}000 lines of CUDA code.  
The system is built on PyTorch~2.3.0 and Python~3.12, running on Ubuntu~22.04 with CUDA~12.1.  
To support large KV cache storage, \textsc{KVDrive} employs \texttt{numpy.memmap} for memory-mapped arrays, enabling persistent and page-level access to KV tensors directly on disk.  
Sparse updates and retrievals are performed through \texttt{torch.Tensor.index\_copy\_()}.  
For clustering, we adopt the Triton kernels from RetroInfer~\cite{retroinfer}, while the data movement primitives for gather and copy are derived from ShadowKV~\cite{sun2024shadowkv}.  
We also leverage FlashInfer~\cite{ye2025flashinfer} for high-performance GPU kernels (such as attention and normalization) optimized for LLM inference. Crucially, our implementation fully supports parallel sessions via continuous batching, a technique widely adopted in modern serving systems like Orca~\cite{yu2022orca} and vLLM~\cite{pagedattention}.
To ensure fair comparisons, all baseline systems mentioned in \autoref{sec:experiment} have been reimplemented and integrated into our unified evaluation framework to ensure consistent configurations and fair comparisons.  
In addition, all long-context models employ the official RoPE implementations provided by the huggingface's \texttt{transformers} framework (e.g., YaRN for Qwen models and LongRoPE for Phi models) to ensure compatibility and consistency across baselines.

\section{Experiment}
\label{sec:experiment}

\subsection{Experimental Setup}
\label{sec:setup}

\noindent
\textbf{\textit{Models.}}
For evaluation, we select four widely used open-source LLMs with strong long-context capabilities:  
Llama-3-8B-1048K~\cite{gradient_llama_3_8b_instruct} (8B parameters, 1M-token context window),  
Qwen3-8B and Qwen3-14B~\cite{qwen3} (128K-token context window),  
and Microsoft Phi-4-mini-instruct~\cite{abdin2024phi4technicalreport} (3.8B parameters, 128K-token context window).  
These models collectively cover a broad spectrum of parameter scales and context lengths, providing a representative basis for evaluating the effectiveness of \textsc{KVDrive} and the baselines.

\noindent
\textbf{\textit{Benchmarks.}} We evaluate \textsc{KVDrive} on two widely-used long-context benchmarks: LongBench~\cite{longbench}, a bilingual dataset for long-context understanding, and RULER~\cite{hsieh2024ruler}, which covers retrieval, multi-hop reasoning, aggregation, and QA tasks.

\noindent
\textbf{\textit{Baselines.}} We compare \textsc{KVDrive} against eight baseline configurations. These include Original, which maintains the entire KV cache in GPU memory without offloading, and FlexGen~\cite{flexgen}, which employs a full-cache strategy by repeatedly loading and evicting KV entries between the GPU and host memory during attention computation. Furthermore, we compare against five state-of-the-art KV cache offloading systems: Quest~\cite{tang2406quest}, ShadowKV~\cite{sun2024shadowkv}, PQCache~\cite{pqcache}, MagicPIG~\cite{chen2025magicpig}, and RetroInfer~\cite{retroinfer}. For a fair comparison, we evaluate two variants of RetroInfer: the original RetroInfer(E) with its native attention estimation enabled, and a modified RetroInfer with this mechanism disabled. For all systems except FlexGen and Original, we apply exact prefilling followed by dynamic sparse attention during the decoding phase. Following common practices in existing systems~\cite{chen2025magicpig,xiao2024efficient}, all sparse attention baselines retain sink tokens (the first 4 tokens) and 64 local tokens in GPU memory, as these consistently exhibit high importance. Finally, for ShadowKV, we adopt the configuration from its original paper, setting the chunk size to 8 and the number of outliers to 48.

\noindent
\textbf{\textit{Hardware Setting.}}
We conduct experiments on three representative hardware environments covering different deployment tiers:  
1)~\textbf{Cost-efficient server:} equipped with an NVIDIA L20 (48~GB) GPU, Intel(R) Xeon(R) Platinum~8457C CPU, and 100~GB DDR5 host memory.  
This configuration represents a practical inference-oriented data-center node with moderate GPU memory and high PCIe bandwidth.  
2)~\textbf{High-end server:} featuring an NVIDIA H20 (96~GB) GPU, AMD EPYC~9K84 96-Core Processor, and 200~GB DDR5 host memory.
It reflects a compute-rich environment typical of enterprise or cloud-scale deployments with larger GPU memory and stronger host-side compute.  
3)~\textbf{Workstation:} powered by an NVIDIA RTX~4090 (24~GB) GPU, Intel(R) Xeon(R) Gold 6430 CPU, and 120~GB DDR5 host memory.
This setup models a resource-constrained local inference scenario representative of edge or on-premise deployments.
The disks in the system are NVMe U.2 SSDs.

\begin{figure*}[t]
	\centering
    \subfloat[][Llama-3-8B-1048K]{
		\begin{minipage}[t]{0.315\linewidth}
		\centering
		\includegraphics[width=\linewidth]{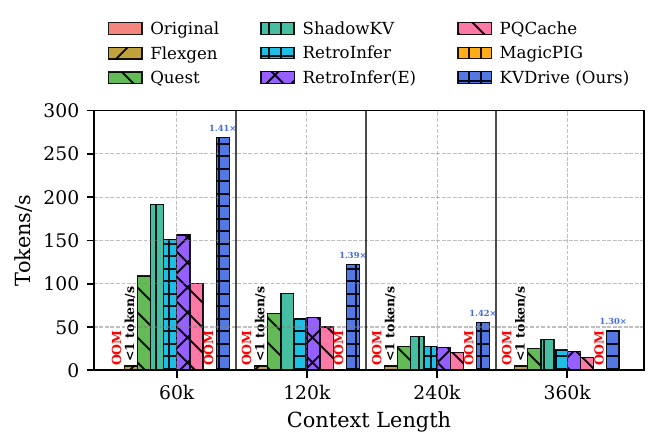}
		\end{minipage}%
		\label{fig:throughput_llama_L20}
	}\hfill
	\subfloat[][Qwen-3-8B]{
		\begin{minipage}[t]{0.315\linewidth}
			\centering
			\includegraphics[width=\linewidth]{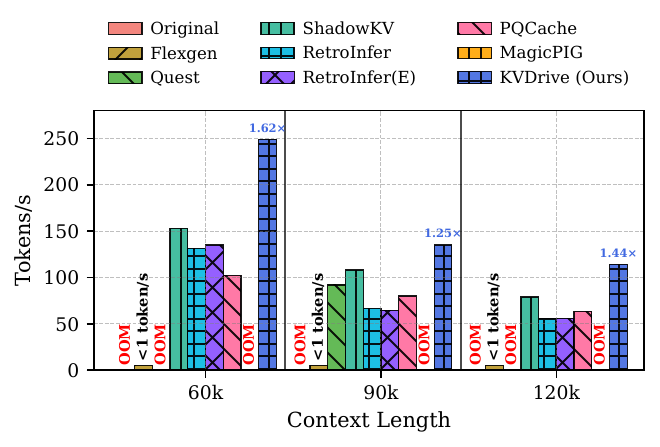}
		\end{minipage}
		\label{fig:throughput_qwen_L20}
	}\hfill
    \subfloat[][Phi-4-Mini-128K]{
		\begin{minipage}[t]{0.315\linewidth}
		\centering
		\includegraphics[width=\linewidth]{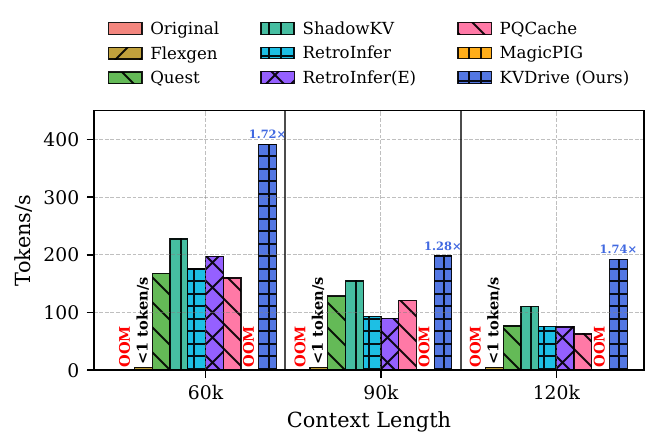}
		\end{minipage}%
		\label{fig:throughput_phi_L20}
	}
    \centering
    \caption{Generation throughput (tokens/s) under varying context lengths and batch sizes in the L20 server. The experiments use batch sizes of 8, 4, 4, 2 and 2 for context lengths of 60k, 90k, 120k, 240k and 360k, respectively.}
    \label{fig:overall_throughput}
\end{figure*}

\begin{figure}[t]
	\centering
    \subfloat[][Qwen-3-14B in H20]{
		\begin{minipage}[t]{0.4\linewidth}
		\centering
		\includegraphics[width=\linewidth]{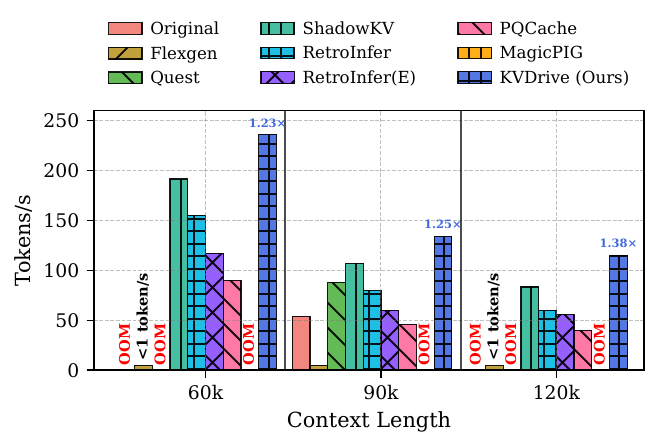}
		\end{minipage}%
		\label{fig:throughput_qwen_H20}
	}
	\subfloat[][Phi-4-Mini-128K in RTX4090]{
		\begin{minipage}[t]{0.4\linewidth}
			\centering
			\includegraphics[width=\linewidth]{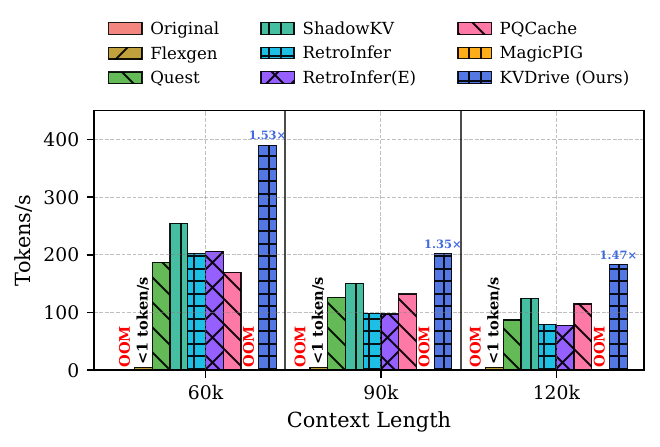}
		\end{minipage}
		\label{fig:throughput_phi_4090}
	}
    \centering
    \caption{Generation throughput (tokens/s) under varying batch sizes and context lengths.}
    \label{fig:overall_throughput_other_server}
\end{figure}

\subsection{Overall Improvement}

\autoref{fig:overall_throughput} presents the generation throughput (in tokens/s) of various systems evaluated across different batch sizes and context lengths using Llama-3-8B-1048K, Qwen-3-8B, and Phi-4-Mini-128K models on an L20 server. \textsc{KVDrive} consistently outperforms all baseline systems across all evaluated configurations. Notably, compared to ShadowKV—the most competitive baseline in our evaluation—\textsc{KVDrive} achieves up to a 70\% improvement in throughput. This significant performance gain is primarily attributed to our proposed elastic pipeline scheduling and efficient cache management mechanisms.
Conversely, several baseline systems struggle to handle the stringent memory and computational demands of long-context generation. For example, MagicPIG fails during initialization because it relies on a massive pre-computed index structure; the memory overhead required to store this index for long-context models exceeds the available host memory capacity, resulting in immediate out-of-memory (OOM) exceptions. Similarly, the Original baseline suffers from OOM failures under heavy workloads, as maintaining a massive KV cache entirely in memory imposes unsustainable memory pressure. Furthermore, among the systems capable of successful initialization, FlexGen demonstrates sub-optimal performance in practical settings. Although it circumvents OOM errors by aggressively offloading data to host memory, its execution model necessitates loading the entire KV cache from off-chip storage during every generation step. This heavily I/O-bound approach incurs severe latency penalties, causing the throughput to plummet to less than 1 token/s. Consequently, such prohibitive performance degradation renders FlexGen impractical for real-world deployment, especially for latency-sensitive applications.

In contrast, \textsc{KVDrive} demonstrates robust and consistent performance across all evaluated hardware configurations. As depicted in \autoref{fig:overall_throughput_other_server}, it achieves throughput improvements ranging from $1.23\times$ to $1.53\times$ over the best-performing baseline on both the H20 and RTX 4090 servers. These results underscore the efficiency and scalability of \textsc{KVDrive}, particularly in scenarios requiring high throughput for both long-context and large-batch inference. Its ability to maintain superior performance under diverse hardware and workload conditions further validates the versatility and robustness of its design.

\begin{table*}[t]
    \setlength{\tabcolsep}{2pt}
    \caption{{
        Performance comparison of different models and methods on RULER and LongBench. 
        }
    }
    \label{T2Iquality}
    \centering
    \scalebox{0.55}{
    \begin{tabular}{cc|cccccccccccccc|cccccccccc}
    \toprule
     & & \multicolumn{14}{c}{RULER} & \multicolumn{10}{c}{LongBench} \\
    \cmidrule(lr){3-16} \cmidrule(lr){17-26}
    Model & Method & S1 & S2 & S3 & MK1 & MK2 & MK3 & MQ & MV & QA-1 & QA-2 & VT & FWE & CWE & Avg. & NQA & MQA & HQA & MQue & DRead & Grep & SAM & PRer & LCC & Avg. \\
    
    \midrule
     \multirow{6}{*}{\makecell{Llama-3\\-8B\\-1048K}} & Full & 100 & 100 & 100 & 98.95 & 97.91 & 43.75 & 98.95 & 96.35 & 75.00 & 48.95 & 78.12 & 72.56 & 0.20  & 77.74 & 18.98 & 41.84 & 36.79 & 21.47 & 31.93 & 34.18 & 35.96 & 81.5 & 56.07 & 39.85 \\
     & Quest & 100&	100&  100&	98.86&	85.42&  21.90&	97.92&	95.49&	70.83&	46.88&	78.75&	65.63&  0.31&	73.99&	17.26&	39.51&	36.78&	18.71&	26.41&	29.49&	35.8&	79.5&	60.05&	38.17  \\
    & ShadowKV & 100&	100&	100&	97.91&	98.95&  32.29 &	96.09&	95.83&	71.87&	52.08&	82.7&	72.56&  0.10&	76.95&	17.17&	39.73&	38.29&	21.08&	31.77&	31.62&	35.87&	80&	63.93&	39.94  \\
    & RetroInfer & 100&	100& 100 &	98.95&	93.75&  37.50&	97.39&	91.66&	73.95&	50.00&	78.54&	63.88&  0.10&	75.82& 19.17& 41.35& 37.59& 19.71& 27.36& 33.76& 43.03& 80.50& 50.77& 39.24	 \\
    & {RetroInfer(E)} & {100}&	{100}& {100} &	{98.95}&	{96.87}&  {41.66}&	{97.91}&	{90.62}&	{73.95}&	{48.95}&	{76.25}&	{73.61}&  {0.10}&	    {76.83}& {18.56}&    {42.55}& {36.89}&    {20.29}& {27.53}&    {33.25}& {42.99}&    {80.00}& {45.38}&    {38.60}	 \\
    & {PQCache} & {85.40}&	{94.80}&     {94.80} &	{92.70}&	{85.41}&  {34.37}&	{95.83}&	{95.05}&	{59.37}&	{43.80}&	{58.12}&	{63.54}&    {0.22}&	{69.49}&    {19.13}& {40.99}&    {36.30}& {20.65}&    {26.21}& {31.77}&    {41.56}& {80.50}&    {44.92}& {38.00}	 \\
    
    & {MagicPIG} & {100}&	{95.83}&     {94.79} &	{86.46}&	{87.50}&  {23.96}&	{78.65}&	{73.96}&	{61.46}&	{44.79}&	{61.46}&	{69.10}&    {0.10}&	{67.54}&    {17.83}& {40.95}&    {37.09}& {21.13}&    {23.81}& {28.98}&    {41.76}& {80.00}&    {43.55}& {37.23}	 \\
    & \cellcolor{gray}\textsc{KVDrive} & \cellcolor{gray} 100& \cellcolor{gray} 100&  \cellcolor{gray} 100& \cellcolor{gray} 98.95& \cellcolor{gray} 94.79& \cellcolor{gray} 37.50& \cellcolor{gray} 98.43& \cellcolor{gray} 94.27& 
    \cellcolor{gray} 71.90& \cellcolor{gray} 51.00& \cellcolor{gray} 75.41& \cellcolor{gray} 67.70& \cellcolor{gray} 0.10& \cellcolor{gray} 76.15&
    \cellcolor{gray} 19.17& \cellcolor{gray} 41.65& 
    \cellcolor{gray} 36.34& \cellcolor{gray} 21.27& \cellcolor{gray} 26.00& \cellcolor{gray} 32.64& \cellcolor{gray} 43.10& \cellcolor{gray} 81.00&
    \cellcolor{gray} 49.84& \cellcolor{gray} 39.00
    \\
    
    \cmidrule{1-26}
    \multirow{6}{*}{\makecell{Qwen-3\\-8B\\}} & Full
    &100 & 100 & 100& 79.16 &62.50 &13.54&95.57&96.09 &59.37  &32.29  &95.83& 83.33& 43.54 & 73.94 & 3.53 & 24.60 & 11.39 & 6.39 & 19.97 & 25.82 & 44.45 & 85.08&65.86 &31.89\\
    
     & Quest & 100 & 98.95 &  98.95 & 83.33 & 44.79 & 1.04 & 91.40 & 97.91 & 41.66 & 35.40 &97.29 & 75.00 & 22.91 & 68.35 & 2.93 & 25.36 & 10.96 & 5.93&20.74 & 27.83&45.88 &85.79 &69.67 &32.79\\
    
    & ShadowKV &100 & 98.95 & 100 & 83.33 & 38.54 & 2.08& 89.32 & 93.48 & 45.83 & 31.25 & 92.08 & 75.69 & 20.93& 67.03 & 2.93 & 23.51 & 10.09 & 6.30 & 19.49 & 27.28 & 45.37 & 86.28 & 69.03 &32.39 \\
    
    & RetroInfer & 100 &  96.90&  98.95&  73.95& 52.08&  2.08&  91.40&  85.40&  34.37&  21.87&  97.70& 74.65&  36.97& 66.64& 2.93 & 26.20 & 11.08 & 5.82 & 19.86 & 27.63 & 46.87 & 87.33 & 68.42 &32.82 \\
    & {RetroInfer(E)} & {100}&	{94.79}& {98.95} &	{75.00}&	{51.04}&  {3.12}&	{90.62}&	{85.67}&	{66.66}&	{46.87}&	{87.91}&	{77.08}&  {16.35}&	    {68.77}& {3.57}&    {25.43}& {11.11}&    {6.27}& {19.85}&    {27.49}& {45.81}&    {86.29}& {67.98}&    {32.64}	 \\
    & {PQCache} & {65.62}&	{79.16}&     {84.37} &	{77.08}&	{30.20}&  {1.04}&	{88.28}&	{93.22}&	{38.54}&	{31.25}&	{82.70}&	{72.56}&    {20.62}&	{58.81}&    {3.51}& {25.56}&    {11.42}& {5.85}&    {22.57}& {27.21}&    {46.06}& {76.45}&    {70.04}& {32.08}	 \\
    
    & {MagicPIG} & {100}&	{83.33}&     {79.16} &	{68.75}&	{34.37}&  {0}&	{66.14}&	{80.20}&	{27.08}&	{21.87}&	{29.58}&	{75.00}&    {4.37}&	{51.52}&    {2.88}& {18.63}&    {10.23}& {5.30}&    {17.81}& {18.75}&    {39.06}& {96.73}&    {56.97}& {29.59}	 \\
    % &   \\
    & \cellcolor{gray}\textsc{KVDrive} & \cellcolor{gray} 100& \cellcolor{gray} 100&  \cellcolor{gray} 100& \cellcolor{gray} 83.33& \cellcolor{gray} 44.79& \cellcolor{gray} 2.08& \cellcolor{gray} 91.92& \cellcolor{gray} 96.35& 
    \cellcolor{gray} 41.66& \cellcolor{gray} 30.20& \cellcolor{gray} 97.29& \cellcolor{gray} 71.52& \cellcolor{gray} 25.83& \cellcolor{gray} 68.07&
    \cellcolor{gray} 3.33& \cellcolor{gray} 24.03& 
    \cellcolor{gray} 10.69& \cellcolor{gray} 6.06& \cellcolor{gray} 19.49& \cellcolor{gray} 27.28& \cellcolor{gray} 45.37& \cellcolor{gray} 86.28&
    \cellcolor{gray} 69.03& \cellcolor{gray} 32.39
    \\
    \cmidrule{1-26}
    \multirow{6}{*}{\makecell{Phi-4\\-Mini\\-128K}} & Full
    &100 & 97.91& 98.95 &83.33 &64.58 &4.16 &81.51 &76.04  &51.04  &38.54 & 73.95 &68.05 & 3.02 & 64.69 & 26.12 & 55.48 & 53.11 & 28.94 & 29.78 & 34.09 & 44.36&91.00 &54.24 &46.34\\
    
     & Quest & 100 & 93.75 &  79.16 & 75.00 & 55.20 & 0 & 76.30 & 72.65 & 45.83 & 38.54 &81.66 & 54.16 & 2.08 & 59.57 & 25.69 & 52.88 & 52.83 & 29.24&29.31 & 33.94&45.13&91.50 &56.19 &46.30\\
    
    & ShadowKV &100 & 95.83 & 68.75 & 79.16 & 53.12 & 1.04& 69.53 & 62.50 & 46.87 & 37.50 & 79.16 & 61.80 & 1.56& 58.21 & 25.69 & 52.63 & 52.89 & 29.49 & 29.92 & 33.06 & 45.41 & 92.00 & 56.99 &46.45 \\
    
    & RetroInfer &100 & 94.79 & 93.75 & 78.12 & 54.16 & 0& 72.39 & 54.42 & 50.00 & 36.45 & 81.45 & 54.51 & 1.66& 59.36 & 25.69& 53.18& 52.93& 28.54&31.45& 33.43& 45.86& 91.50& 56.13& 46.52\\

     & {RetroInfer(E)} & {97.91}&	{73.95}& {94.79} &	{54.16}&	{37.50}&  {0}&	{47.39}&	{36.71}&	{30.20}&	{31.25}&	{65.62}&	{56.94}&  {1.45}&	    {48.29}& {26.33}&    {48.09}& {46.49}&    {24.36}& {29.02}&    {31.36}& {42.06}&    {91.50}& {43.24}&    {42.49}	 \\
    & {PQCache} & {35.41}&	{3.12}&     {2.08} &	{6.25}&	{6.25}&  
    {0}&	{4.16}&	
    {6.25}&	{19.79}&	{26.04}&	{38.12}&	{57.98}&    {0.41}&	{15.83}&    {26.14}& {47.60}&    {45.23}& {21.13}&    {28.98}& {29.34}&    {41.36}& {91.00}&    {40.55}& {41.25}	 \\
    
    & {MagicPIG} & {50}&	{14.58}&     {0} &	{9.37}&	{16.66}&  {0}&	{2.86}&
    {1.56}&	{20.83}&	{29.16}&	{19.79}&	{47.56}&    {0.52}&	{16.09}&    {22.86}& {37.96}&    {43.11}& {20.41}&    {17.55}& {15.19}&    {35.28}& {88.02}&    {32.59}& {34.77}	 \\
    
    % &   \\
    & \cellcolor{gray}\textsc{KVDrive} & \cellcolor{gray} 100& \cellcolor{gray} 94.79&  \cellcolor{gray} 90.62& \cellcolor{gray} 79.16& \cellcolor{gray} 52.10& \cellcolor{gray} 1.04& \cellcolor{gray} 71.61& \cellcolor{gray} 60.67& 
    \cellcolor{gray} 48.95& \cellcolor{gray} 36.45& \cellcolor{gray} 82.70& \cellcolor{gray} 59.72& \cellcolor{gray} 1.14& \cellcolor{gray} 59.91&
    \cellcolor{gray} 25.69& \cellcolor{gray} 52.24& 
    \cellcolor{gray} 53.16& \cellcolor{gray} 28.95& \cellcolor{gray} 29.69& \cellcolor{gray} 32.84& \cellcolor{gray} 45.05& \cellcolor{gray} 91.50&
    \cellcolor{gray} 53.79& \cellcolor{gray} 45.87
    \\
    % \midrule
    \bottomrule
    \end{tabular}}
\end{table*}

\begin{table}[t]
\setlength{\tabcolsep}{3.4pt}
\caption{Effectiveness of the Lookahead (LA) eviction policy across different systems.}
\label{tab:hit_rate}
\centering
\scalebox{0.65}{
\begin{tabular}{l|ccc|ccc|ccc}
\toprule
\multirow{2}{*}{Method} & \multicolumn{3}{c|}{Llama3-8B-1048K} & \multicolumn{3}{c|}{Qwen-3-8B} & \multicolumn{3}{c}{Phi-4-mini-128K} \\
\cmidrule{2-10}
 & LRU(\%) & LA(\%) & Imp.(\%) & LRU(\%) & LA(\%) & Imp.(\%) & LRU(\%) & LA(\%) & Imp.(\%) \\
\midrule
Quest     & 89.2 & 90.1 & +0.9  & 89.7 & 91.0 & +1.3  & 90.7 & 88.9 & -1.8 \\
ShadowKV  & 84.2 & 87.3 & +2.9  & 81.0 & 83.2 & +2.2  & 86.1 & 84.6 & -1.5 \\
RetroInfer     & 80.7 & 82.0 & +1.3  & 70.0 & 73.9 & +3.9  & 78.3 & 80.1 & +1.9  \\
\textsc{KVDrive}   & 80.0 & 81.5 & +1.5  & 77.5 & 79.0 & +1.5  & 78.0 & 81.0 & +3.0  \\
\bottomrule
\end{tabular}
}
\end{table}

\subsection{Micro Benchmark}

\noindent
\textbf{\textit{Accuracy}}.
We first examine whether the system-level optimizations in \textsc{KVDrive} compromise model accuracy. 
\autoref{T2Iquality} reports the results on both RULER and LongBench. 
Across all tasks, \textsc{KVDrive} achieves accuracy that is on par with or even slightly better than state-of-the-art offloading systems such as Quest, ShadowKV, and InfiniGen. 
This demonstrates that our indexing and retrieval design in \autoref{sec:system_overview} introduces negligible accuracy loss, even under long-context (up to 128K tokens). 
The results confirm that the gains of \textsc{KVDrive} primarily stem from improved cache management and scheduling, rather than trading accuracy for efficiency.

\begin{figure}[t]
    \centering
    \includegraphics[width=0.7\linewidth]{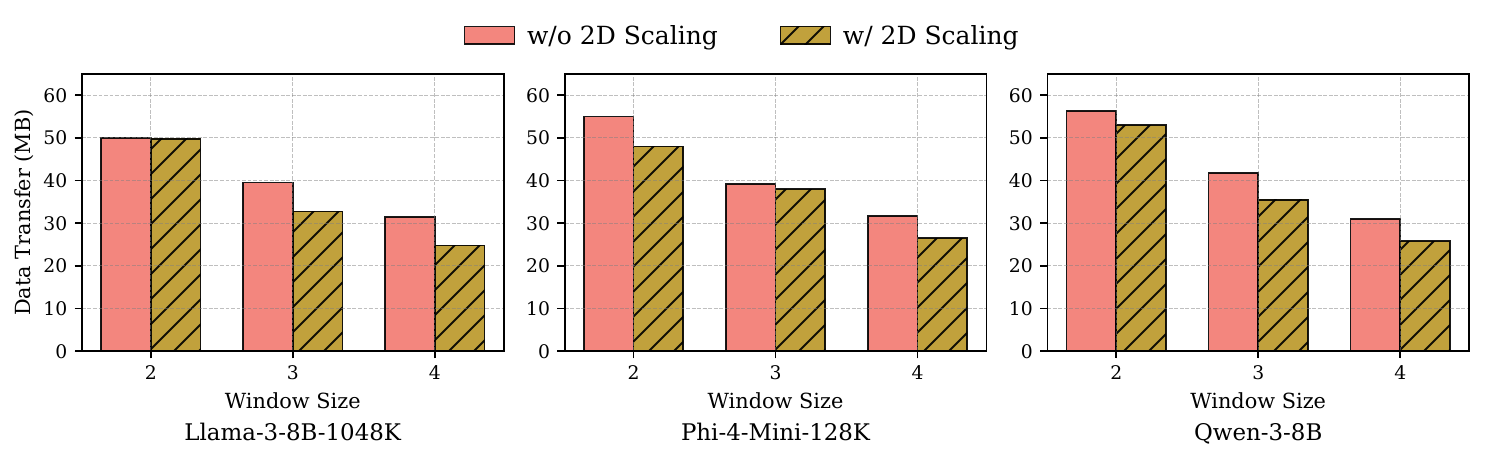}
    \caption{Impact of 2D window scaling on data transfer volume under different window sizes and models.}
    \label{fig:2D_scaling_IO}
\end{figure}

\begin{figure}[t]
    \centering
    \includegraphics[width=0.7\linewidth]{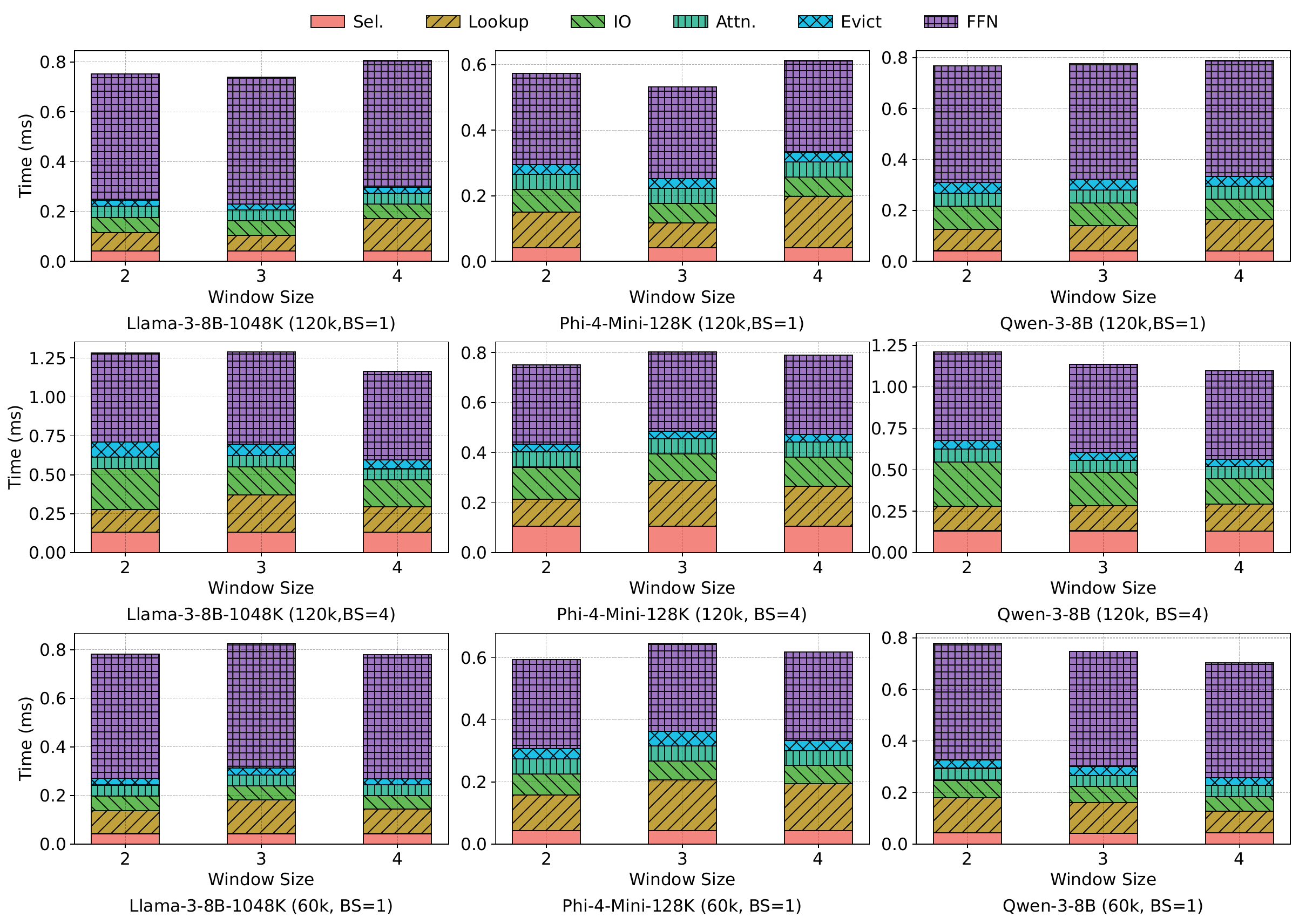}
    \caption{Time breakdown of \textsc{KVDrive} under different window sizes and models at 1.56\% sparsity.}
    \label{fig:time_breakdown_comparison_window_size}
\end{figure}

\noindent
\textbf{\textit{Generality of Eviction Policy.}} \autoref{tab:hit_rate} evaluates the generality of our Lookahead (LA) eviction policy by applying it to four different systems: Quest, ShadowKV, RetroInfer, and \textsc{KVDrive}. The experiments are conducted with a 120k context length, window size of 2, sparsity budget of 2048, and a look-ahead candidate pool $M=2560$. As shown in the table, the LA strategy demonstrates broad applicability, outperforming the traditional LRU baseline in the majority of configurations. Specifically, on Llama3-8B-1048K and Qwen-3-8B, the LA policy consistently boosts hit rates across all evaluated methods, achieving gains ranging from 0.9\% to 3.9\%. This indicates that identifying eviction candidates based on attention scores (as detailed in \S\ref{sec:lookahead_eviction}) is a robust approach for various system architectures. While Quest achieves marginally higher hit rates with both policies, it incurs significant overhead: its index size is over $4\times$ larger than that of \textsc{KVDrive} and $2\times$ larger than ShadowKV. In contrast, \textsc{KVDrive} with the LA policy achieves a comparable hit rate while maintaining a significantly smaller memory footprint.

\noindent
\textbf{\textit{2D Window Scaling.}} 
\autoref{fig:2D_scaling_IO} reports the ablation results of our 2D window scaling strategy for Llama-3-8B-1048K, Phi-4-Mini-128K, and Qwen-3-8B in \autoref{sec:2D_scaling}. 
Under a constrained GPU memory budget, 2D scaling results in reduced data transfer compared to uniform window allocation. This indicates that heterogeneous reuse patterns across layers and attention heads can effectively improve the GPU cache utilization and avoid the waste of GPU cache.

\noindent
\textbf{\textit{Window Size.}} \autoref{fig:time_breakdown_comparison_window_size} depicts the latency breakdown of a single Transformer layer under a 1.56\% sparsity budget. We evaluate the performance by sweeping window sizes through \{2, 3, 4\} across context lengths of 60k and 120k, and batch sizes of 1 and 4. As the window size increases, the reduction in data transfer volume alleviates I/O overhead; however, this benefit is offset by increased lookup latency. This trade-off necessitates a careful balance between I/O efficiency and lookup overhead. Specifically, with a batch size of 1, smaller window sizes (e.g., 2) generally yield lower latency. This is attributed to the surge in lookup time as cache capacity expands, while I/O bandwidth remains underutilized. Conversely, at a batch size of 4, larger window sizes (e.g., 4) prove more effective, primarily due to the reduced demand on I/O bandwidth. These results underscore the critical role of window size selection in maximizing throughput. Furthermore, as the batch size scales from 1 to 4, the latency contribution from non-FFN modules—including selection, lookup, attention, and I/O—increases significantly. This shift indicates that non-FFN components emerge as the dominant bottleneck at larger batch sizes, highlighting the imperative to optimize these modules for scalability.
\begin{figure}[t]
    \centering
    \includegraphics[width=0.7\linewidth]{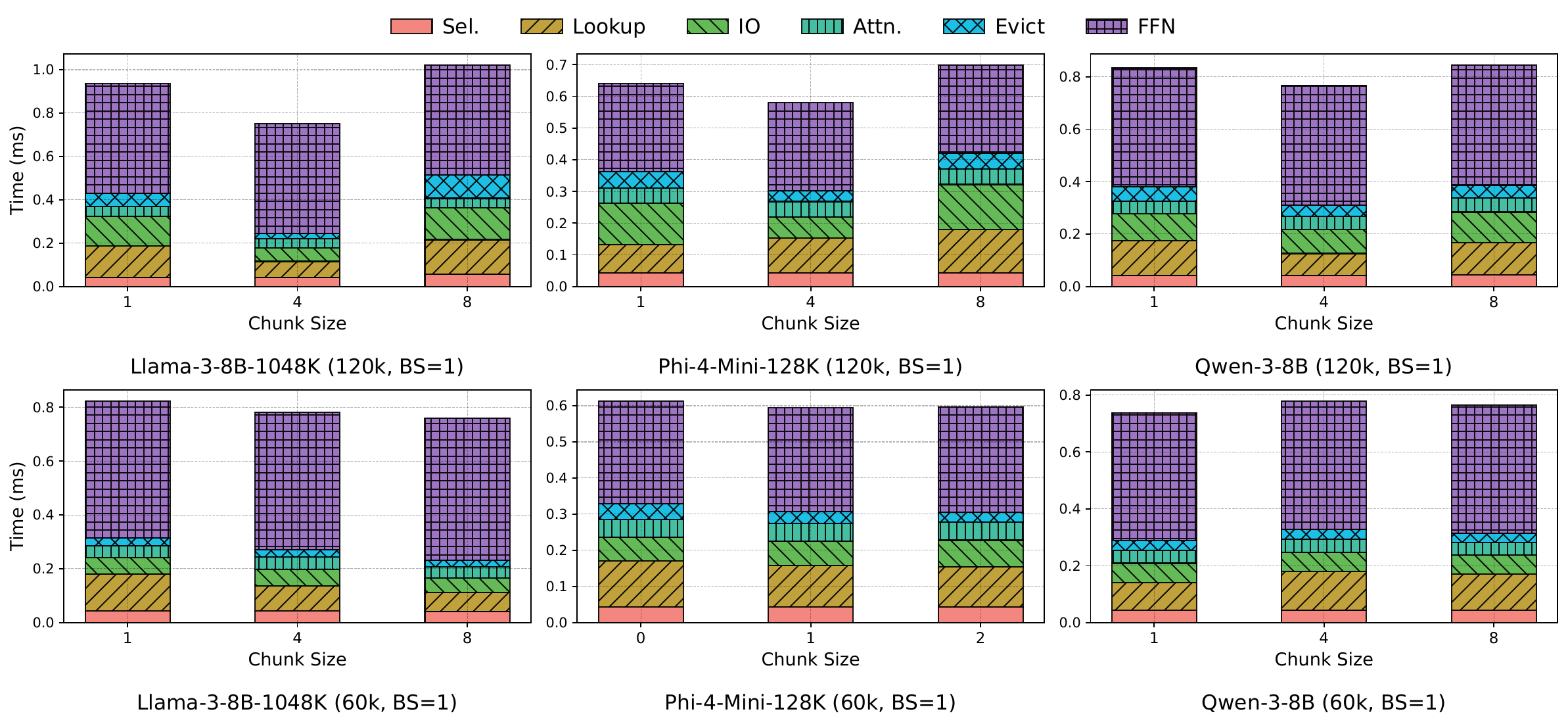}
    \caption{Time breakdown of \textsc{KVDrive} under different chunk sizes at 1.56\% sparsity.}
    \label{fig:time_breakdown_chunk}
\end{figure}

\begin{figure}[t]
    \centering
    \includegraphics[width=0.7\linewidth]{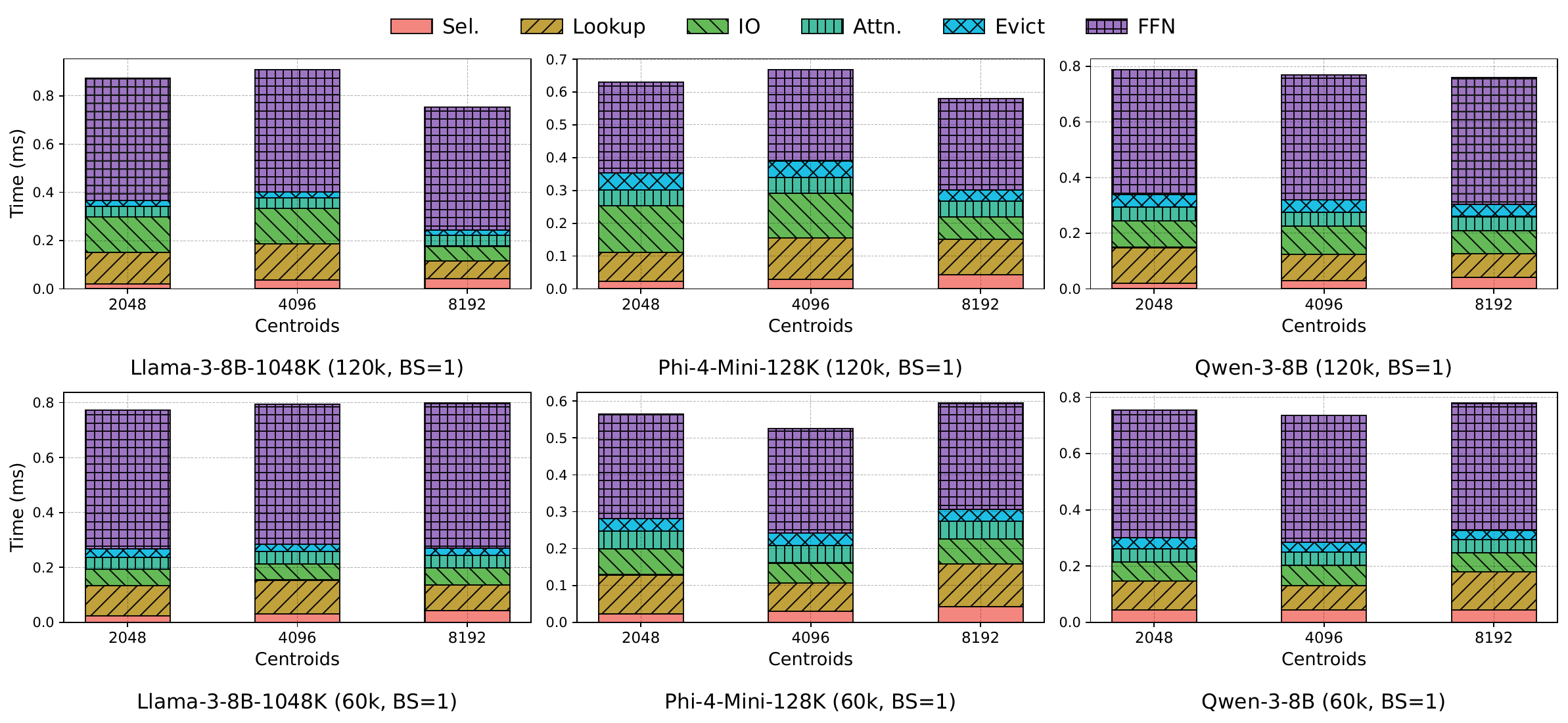}
    \caption{Time breakdown of \textsc{KVDrive} under different numbers of centroids at 1.56\% sparsity.}
    \label{fig:time_breakdown_centroids}
\end{figure}

\noindent
\textbf{\textit{Chunk size.}} \autoref{fig:time_breakdown_chunk} depicts the latency breakdown of a single Transformer layer under a 1.56\% sparsity budget. We evaluate performance across context lengths of 60k and 120k, sweeping chunk sizes through \{1, 4, 8\}. Notably, we observe a distinct U-shaped trend in both overall latency and specific I/O latency as the chunk size increases. This behavior reflects a fundamental trade-off between semantic continuity and data redundancy. With small chunk sizes (e.g., 1), data is processed in fragmented units, necessitating frequent memory accesses that inflate I/O overhead. Conversely, excessively large chunk sizes (e.g., 8) introduce redundant information, which increases the data transfer volume and negates the benefits of reduced access frequency. An intermediate chunk size (e.g., 4) strikes an optimal balance, preserving sufficient semantic continuity while minimizing redundancy, thereby achieving the lowest I/O cost.

\begin{figure}[t]
    \centering
    \includegraphics[width=0.8\linewidth]{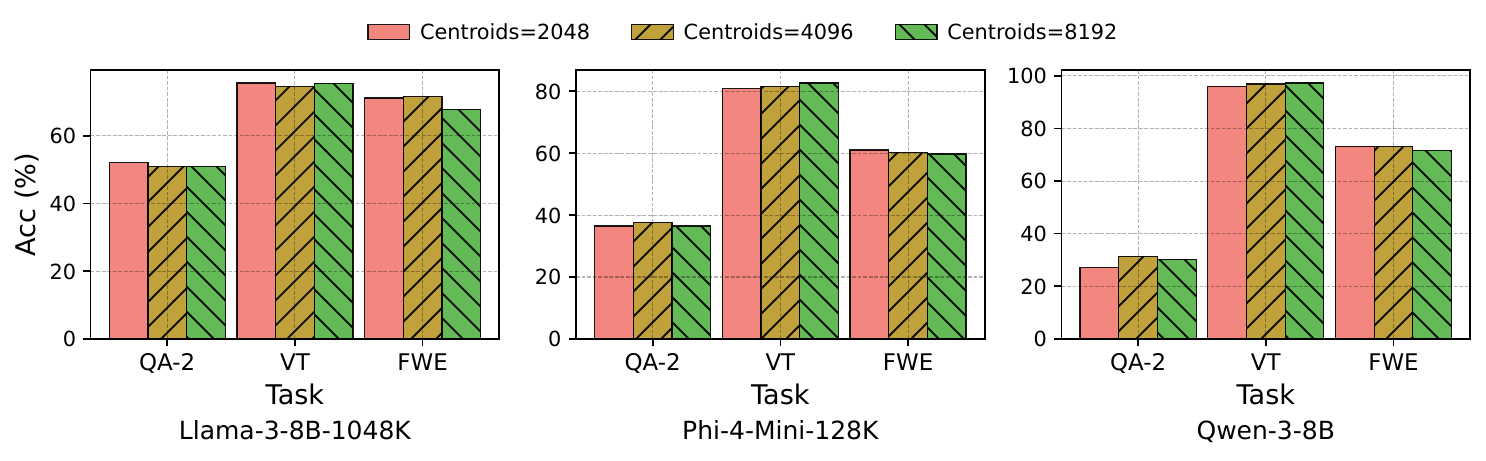}
   \caption{Accuracy under different numbers of centroids across tasks.}
    \label{fig:acc_centroids}
\end{figure}

\begin{figure}[t]
    \centering
    \includegraphics[width=0.8\linewidth]{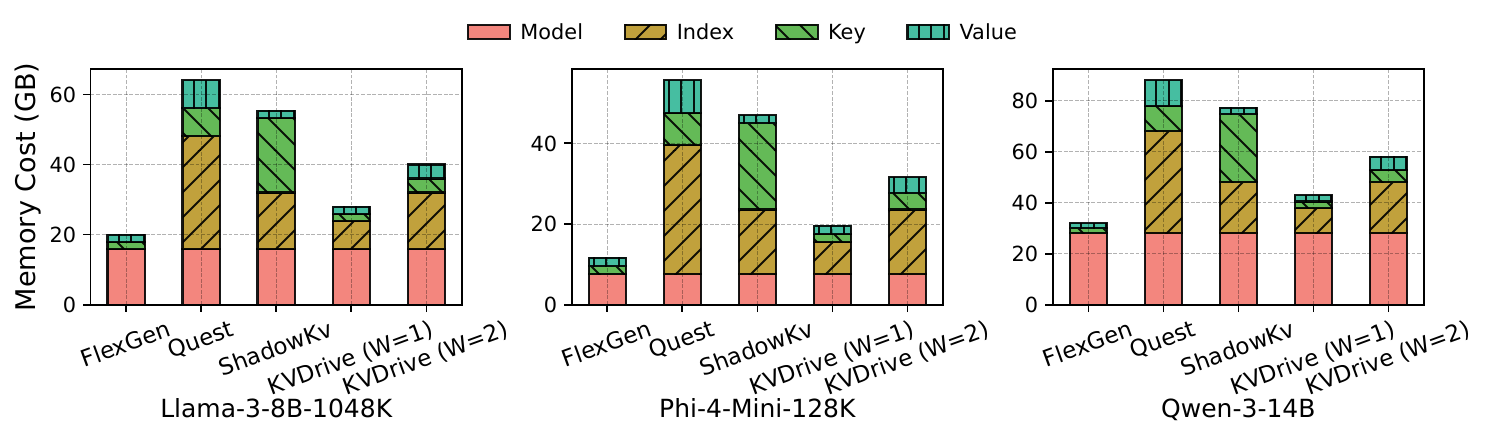}
   \caption{Memory layout comparison across different models.}
    \label{fig:memory_cost_ratio}
\end{figure}

\noindent
\textbf{\textit{Number of Centroids.}}
\autoref{fig:time_breakdown_centroids} decomposes the latency of executing a single Transformer layer. We evaluate the performance across context lengths of 60k and 120k, sweeping the number of centroids through \{2,048, 4,096, 8,192\} under a fixed 1.56\% sparsity budget. We observe a consistent relationship between context size and the optimal number of centroids: specifically, 8,192 centroids yield the best performance for the 120k context, while 4,096 centroids are optimal for 60k. This suggests that the optimal ratio of context length to centroids remains constant. While increasing the number of centroids incurs higher selection latency due to the added computational complexity of clustering, this overhead is effectively mitigated by significant reductions in I/O and lookup times. Higher centroids count facilitate finer-grained clustering, which distributes data more evenly across clusters. This improved load balance reduces synchronization overhead and enhances data locality, thereby lowering the latency of memory-bound components.

\noindent
\textbf{\textit{Impact of Centroid Reduction.}}
\autoref{fig:acc_centroids} examines the accuracy of \textsc{KVDrive} under varying numbers of centroids for tasks such as QA-2, Variable Tracking, and Frequent Words Extraction. The results demonstrate that accuracy remains consistent regardless of the number of centroids, highlighting the robustness of the clustering approach. Notably, reducing the number of centroids from 8192 to 2048 achieves a $4\times$ reduction in index size without any loss in precision. This observation underscores the effectiveness of the clustering mechanism in preserving the semantic structure required for accurate task performance, even with fewer centroids. The reduction in index size directly translates to lower memory and storage overhead, improving the overall efficiency of the method. 

\begin{figure}[t]
	\centering
    \subfloat[][Llama-3-8B-1048K]{
		\begin{minipage}[t]{0.35\linewidth}
		\centering
		\includegraphics[width=\linewidth]{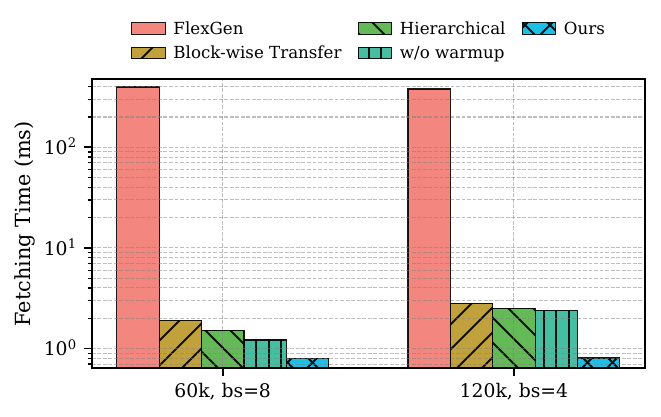}
		\end{minipage}%
		\label{fig:ssd_latency}
	}
	\subfloat[][Llama-3-8B-1048K]{
		\begin{minipage}[t]{0.35\linewidth}
			\centering
			\includegraphics[width=\linewidth]{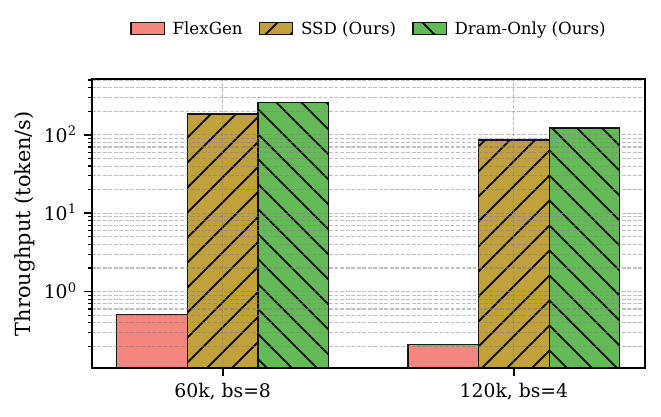}
		\end{minipage}
		\label{fig:ssd_throughput}
	}
    \centering
	\caption{Performance comparison in DRAM-Only and DRAM + SSD.}
    \label{fig:ssd}
\end{figure}

\begin{figure}[t]
    \centering
    \includegraphics[width=0.7\linewidth]{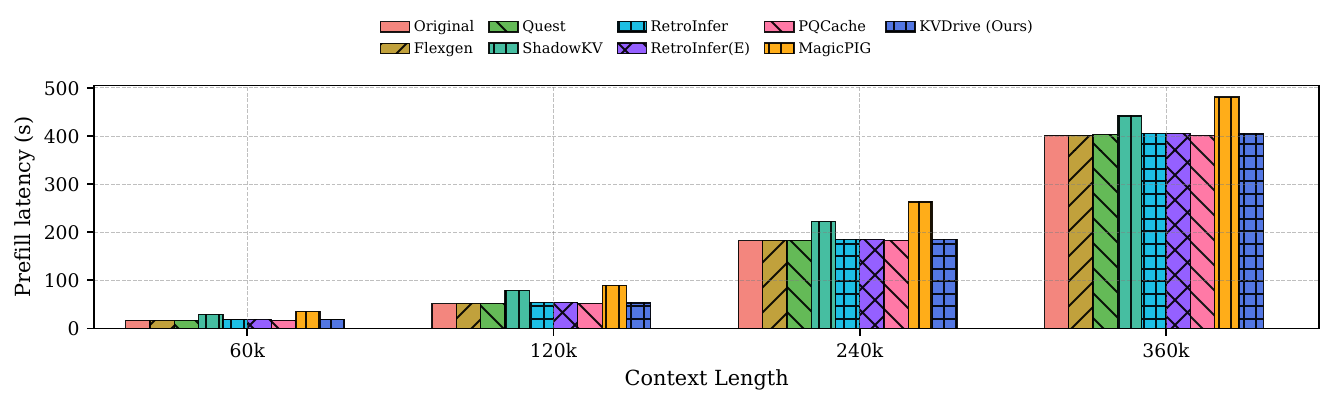}
    \caption{Prefill latency (s) under different context lengths.}
    \label{fig:prefill_latency}
\end{figure}

\noindent
\textbf{\textit{Memory Layout.}} With a batch size of 8 and a context length of 120k, \autoref{fig:memory_cost_ratio} evaluates the GPU memory usage of various systems across three models: Llama-3-8B-1048K, Qwen-3-14B, and Phi-4-Mini-128K. Among the evaluated systems, \textsc{KVDrive} exhibits significantly lower GPU memory usage compared to ShadowKV and Quest. This result is primarily attributed to its sparse and efficient indexing mechanism. Conversely, ShadowKV incurs substantial memory overhead due to its reliance on compressed key storage, which is entirely resident in GPU memory and imposes significant cost. FlexGen achieves the lowest GPU memory utilization by avoiding in-memory storage of indexes. However, this design comes at the expense of performance: FlexGen must reload the entire KV cache for a layer at every generation step, introducing severe I/O bottlenecks and extreme latency. In contrast, \textsc{KVDrive}'s memory footprint, while slightly higher than that of FlexGen, avoids these I/O penalties through its efficient caching and sparse indexing strategies. Moreover, as the window size increases (e.g., from 1 to 2), \textsc{KVDrive} maintains lower GPU memory usage compared to Quest and ShadowKV, while significantly alleviating I/O pressure.

\noindent
\textbf{\textit{DRAM-Only VS DRAM + SSD.}} \autoref{fig:ssd} compares the performance of FlexGen and our approach under various configurations for the Llama-3-8B-1048K model, focusing on fetching time and throughput. As shown in \autoref{fig:ssd_latency}, our Blockwise Transfer strategy delivers substantial performance improvements over FlexGen. By sparsely fetching only the required blocks on demand, our approach significantly reduces fetching overhead compared to FlexGen’s full-layer loading mechanism. Moreover, the Hierarchical strategy with asynchronous fetching further minimizes latency by enabling more efficient lookups and data access. Additionally, the incorporation of a prefill warmup stage, which identifies critical KV cache entries during the final prefill phase, further optimizes fetching time by prioritizing high-value cache components. \autoref{fig:ssd_throughput} demonstrates that our approach achieves significantly higher throughput than FlexGen across all configurations. Notably, when using SSD as a component of the storage hierarchy, \textsc{KVDrive} sustains high throughput with only a 40\% reduction compared to the DRAM-only configuration, while supporting larger batch sizes.

\noindent
\textbf{\textit{Prefill Latency.}} \autoref{fig:prefill_latency}  evaluates the prefill latency of Llama-3-8B-1048K. KVDrive matches the performance of the full-attention baseline (Original) across all context lengths, indicating that its index construction and offloading incur negligible overhead. This efficiency is attributed to the low computational cost of our K-Means clustering—which is asymptotically superior to the quadratic self-attention—and the effective masking of offloading traffic via computation-communication overlapping. In comparison, MagicPIG and ShadowKV exhibit higher latencies due to specific bottlenecks: MagicPIG is limited by large LSH table construction, while ShadowKV incurs extra costs from low-rank key decomposition.

\begin{figure}[t]
	\centering
    \subfloat[][Llama-3-8B-1048K]{
		\begin{minipage}[t]{0.35\linewidth}
		\centering
		\includegraphics[width=\linewidth]{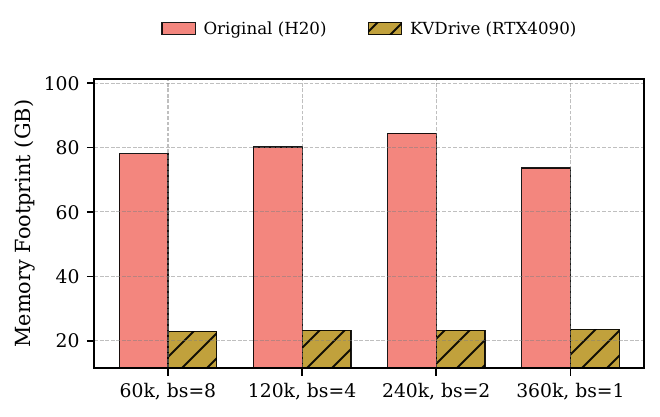}
		\end{minipage}%
		\label{fig:cost_memory}
	}
	\subfloat[][Llama-3-8B-1048K]{
		\begin{minipage}[t]{0.35\linewidth}
			\centering
			\includegraphics[width=\linewidth]{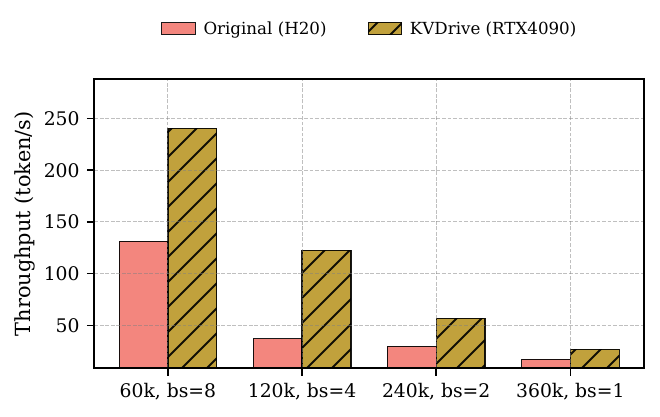}
		\end{minipage}
		\label{fig:cost_throughput}
	}
    \centering
	\caption{Cost-efficiency analysis on Llama-3-8B-1048K.}
    \label{fig:cost_analyse}
\end{figure}

\noindent
\textit{\textbf{Cost-Efficiency Analysis.}} To demonstrate the economic advantages of our design, \autoref{fig:cost_analyse} compares the Llama-3-8B-1048K inference performance on two hardware tiers: a high-end NVIDIA H20 (96\,GB HBM) using standard in-memory serving, and a consumer-grade NVIDIA RTX 4090 (24\,GB HBM) powered by \textsc{KVDrive}. \autoref{fig:cost_memory} shows that \textsc{KVDrive} reduces the memory footprint by approximately $4\times$ through effective sparse offloading, whereas the H20 baseline approaches memory saturation even with its larger capacity. By breaking the memory wall, \textsc{KVDrive} enables the RTX 4090 to achieve up to $3\times$ higher throughput than the H20 baseline (\autoref{fig:cost_throughput}). This result proves that with an optimized storage hierarchy, consumer hardware can effectively service long-context workloads that were previously limited to enterprise-grade GPUs.

\begin{figure}[t]
    \centering
    \includegraphics[width=\linewidth]{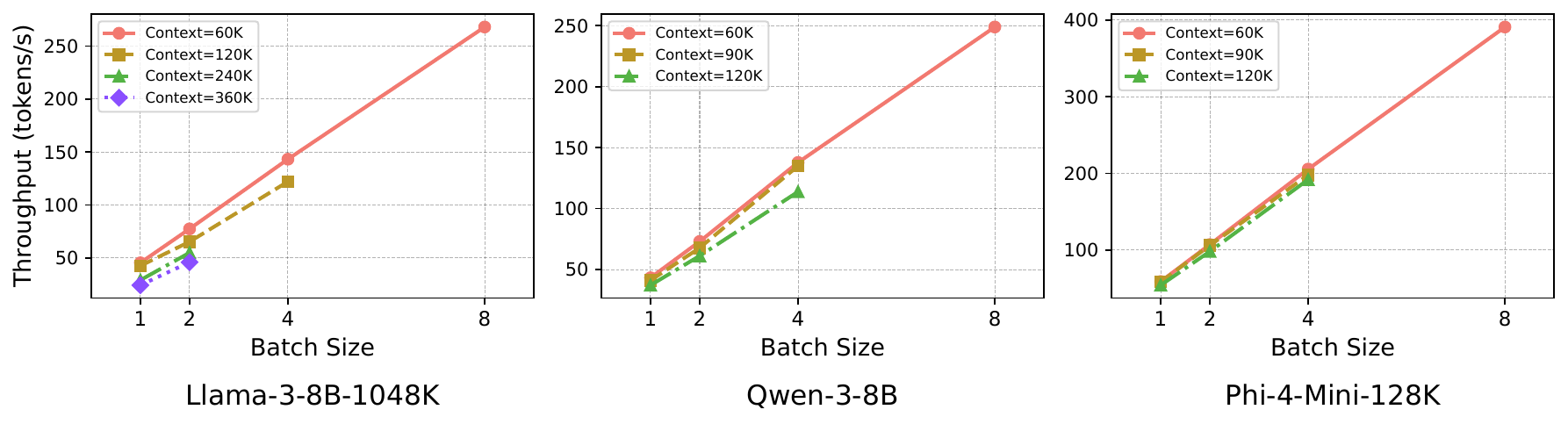}
    \caption{Generation throughput (tokens/s) under different batch sizes.}
    \label{fig:batch_size}
\end{figure}

\noindent
\textbf{Batch Size.} \autoref{fig:batch_size} presents the generation throughput of \textsc{KVDrive} with different batch sizes, evaluated across multiple models and context lengths. As shown, the system maintains a robust upward trend in throughput as batch size scales. This efficiency stems from our execution mechanism, which overlaps I/O and CPU-side pre-processing with GPU-side computation across micro-batches. By interleaving these distinct operations, \textsc{KVDrive} effectively hides data movement overheads and maximizes GPU utilization, thereby supporting larger batch sizes without significant performance degradation.

% \noindent
% \textbf{\textit{Long-Term Memory Capacity.}} We evaluate \textsc{KVDrive} on \textit{LongMemEval-s} with approximately 115k tokens per problem~\cite{longmemeval}.
% ...

\section{Conclusion}

We present \textsc{KVDrive}, a holistic multi-tier KV cache management system for long-context LLM inference. 
\textsc{KVDrive} introduces three system-level techniques: attention-based in-GPU cache management with sliding-window reuse and lookahead eviction, elastic pipeline scheduling that overlaps selection, fetching, and computation with minimized stalls, and coordinated multi-tier KV storage. 
Our evaluation shows that \textsc{KVDrive} improves throughput by 1.74$\times$ over state-of-the-art offloading systems, while preserving accuracy. 
These results demonstrate the effectiveness of system-level cache and pipeline co-design in enabling efficient long-context inference under tight GPU budgets.
% Based on \textsc{KVDrive}, our future work will focus on two key aspects: extending our holistic management to multimodal models, which present distinct KV cache access patterns, and investigating Processing-in-Memory hardware to offload selection and partial computation directly into the storage tiers, further mitigating data movement bottlenecks.

\section{Future Work}

Building on \textsc{KVDrive}, our future research will focus on three key directions. First, we aim to extend our holistic management to multimodal models, which present distinct KV cache access patterns compared to text-only LLMs. Second, we plan to investigate Processing-in-Memory hardware to offload selection and partial computation directly into storage tiers, further mitigating data movement bottlenecks.
Finally, we plan to explore the synergy between \textsc{KVDrive} and compression techniques, such as quantization and pruning. We envision a \textit{Tiered Mixed-Precision Storage} scheme that leverages \textsc{KVDrive}'s ability to distinguish hot and cold KV blocks and assigns precision adaptively across tiers: maintaining high precision (e.g., FP16) for latency-critical hot blocks in HBM, while applying more aggressive quantization (e.g., INT4) to cold blocks spilled to SSD.
This design trades modest (de)quantization overhead for higher effective I/O bandwidth and storage capacity, aiming to alleviate the I/O bottleneck in massive-context retrieval while preserving end-to-end generation quality.

\bibliographystyle{ACM-Reference-Format}
\bibliography{sample-base}

% Articles V4mod124-V4mod249 use

\end{document}